\documentclass[10pt,twocolumn,letterpaper]{article}

\usepackage[pagenumbers]{cvpr} 

\usepackage{url}
\usepackage{graphicx}
\usepackage{amsmath}
\usepackage{amsthm}
\usepackage{amssymb}
\usepackage{booktabs}
\usepackage{bbm}
\usepackage{multicol}
\usepackage{multirow}
\usepackage{graphicx}
\usepackage{xcolor}
\usepackage{enumitem}
\usepackage{siunitx}
\usepackage{colortbl}
\usepackage{etoolbox}
\robustify\cellcolor

\usepackage{graphicx}
\usepackage{float}

\usepackage[pagebackref,breaklinks,colorlinks]{hyperref}

\usepackage[capitalize]{cleveref}
\crefname{section}{Sec.}{Secs.}
\Crefname{section}{Section}{Sections}
\Crefname{table}{Table}{Tables}
\crefname{table}{Tab.}{Tabs.}

\def\cvprPaperID{4593} 
\def\confName{CVPR}
\def\confYear{2023}

\newcommand{\yulin}[1]{{\color{black}#1}}

\newcommand{\haoran}[1]{{\color{black}#1}}

\newcommand{\SO}{\mathrm{SO}(3)}
\newcommand{\Det}[1]{\operatorname{det}(#1)}

\graphicspath{image/}

\usepackage{booktabs}

\begin{document}

\title{Delving into Discrete Normalizing Flows on SO(3)\\
Manifold for Probabilistic Rotation Modeling}


\author{
Yulin Liu\thanks{Equal contribution} \hskip 1.6em
Haoran Liu\footnotemark[1] \hskip 1.6em
Yingda Yin\footnotemark[1] \hskip 1.6em 
Yang Wang \hskip 1.6em 
Baoquan Chen\thanks{He Wang and Baoquan Chen are the corresponding authors
(\{hewang, baoquan\}@pku.edu.cn).
}  \hskip 1.6em 
He Wang\footnotemark[2]  \\
Peking University \\
}

\maketitle
\begin{abstract}
Normalizing flows (NFs) provide a powerful tool to construct an expressive distribution by a sequence of trackable transformations of a base distribution and form a probabilistic model of underlying data.
Rotation, as an important quantity in computer vision, graphics, and robotics, can exhibit many ambiguities when occlusion and symmetry occur and thus demands such probabilistic models.  
Though much progress has been made for NFs in Euclidean space,
there are no effective normalizing flows without discontinuity or many-to-one mapping tailored for $\SO$ manifold. Given the unique non-Euclidean properties of the rotation manifold, adapting the existing NFs to $\SO$ manifold is non-trivial. 
In this paper, we propose a novel normalizing flow on $\SO$ by combining a Mobius transformation-based coupling layer and a quaternion affine transformation.
With our proposed rotation normalizing flows, one can not only effectively express arbitrary distributions on $\SO$, but also conditionally build the target distribution given input observations.
Extensive experiments show that our rotation normalizing flows significantly outperform the baselines on both unconditional and conditional tasks.

\end{abstract}

\section{Introduction}

Endowing a neural network with the ability to express uncertainty along with the prediction is of crucial influence to safety and interpretability-critical systems and provides valuable information for downstream tasks \cite{schwalbe2020survey,leibig2017leveraging,ching2018opportunities}. As a widely used technique in computer vision and robotics, rotation regression can also benefit from such uncertainty-aware predictions and enable many applications \cite{sattler2019understanding,glover2012monte,crassidis2003unscented}. 

To this end, recent years have witnessed much effort in modeling the uncertainty of rotation via probabilistic modeling of the $\SO$ space, including von Mises distribution for Euler angles \cite{prokudin2018deep}, Bingham distribution for quaternions \cite{gilitschenski2019deep,deng2022deep}, matrix Fisher distribution for rotation matrices \cite{mohlin2020probabilistic}, etc. Those distributions are all single-modal, which fall short on modeling objects with continuous symmetry, which are ubiquitous in our daily life. Taking \textit{cup} as an example, it exhibits rotational symmetry for which modeling with the unimodal or the mixture of distributions is clearly insufficient. How to model an \textit{arbitrary} distribution on $\SO$ manifold is still a challenging open problem.

Normalizing flows \cite{rezende2015variational}, which maps samples from a simple base distribution to the target distributions via invertible transformations, provides a flexible way to express complex distributions and has been widely used in expressing arbitrary distributions in Euclidean space \cite{dinh2014nice,dinh2016density,kingma2018glow,ho2019flow++,behrmann2019invertible,chen2019residual}. However, developing normalizing flows on $\SO$ manifold is still highly under-explored.

Some works rely on normalizing flows in Euclidean space and adapt them to handle rotations. ReLie \cite{falorsi2019reparameterizing} proposes normalizing flows for general Lie group via Lie algebra in Euclidean space. However, it suffers from discontinuous rotation representations \cite{Zhou2019Continuity} and leads to inferior performance. 
ProHMR \cite{kolotouros2021probabilistic} considers rotations as 6D vectors in Euclidean space and leverages Glow \cite{kingma2018glow} to model distributions, where a many-to-one Gram-Schmidt projection is needed to close the gap between the two spaces. Although composed of bijective transformations in Euclidean space, the many-to-one mapping from Euclidean space to $\SO$ breaks the one-to-one regulation of normalizing flows \cite{rezende2015variational}.

Other works propose general normalizing flows for non-Euclidean spaces.
Mathieu et al.\cite{mathieu2020riemannian}, Lou et al.\cite{lou2020neural} and Falorsi et al.\cite{falorsi2020neural} propose continuous normalizing flows for general Riemannian manifold, without considering any property of $\SO$ space, which leads to unsatisfactory performance for probabilistic rotation modeling. 
Rezende et al. \cite{rezende2020normalizing} introduce normalizing flows on tori and spheres. Note that despite unit quaternions lying on $\mathcal{S}^3$ space, \cite{rezende2020normalizing} does not exhibit the antipodal symmetry property of quaternions and thus is not suitable for modeling rotations in $\SO$.

In this work, we introduce novel discrete normalizing flows for rotations on the $\SO$ manifold. The core building block of our discrete rotation normalizing flow consists of a Mobius coupling layer with rotation matrix representation and an affine transformation with quaternion representation, linked by conversions between rotations and quaternions. In the Mobius coupling layer, one column of the rotation matrix acts as the \textit{conditioner}, remaining unchanged. Another column serves as the \textit{transformer}, undergoing Mobius transformation conditioned on the conditioner, while the remaining column is determined by the cross-product. By combining multiple Mobius coupling layers, we enhance the capacity of our vanilla design for rotation normalizing flows.

To further increase expressivity, we propose an affine transformation in quaternion space. This quaternion affine transformation complements the Mobius coupling layer, functioning as both a global rotation and a means for condensing or dilating the local likelihood. Despite quaternions being a double coverage of the $\SO$ manifold, this transformation remains bijective and diffeomorphic to $\SO$.

We conduct extensive experiments to validate the expressivity and stability of our proposed rotation normalizing flows. The results show that our rotation normalizing flows are able to either effectively fit the target distributions on $\SO$ with distinct shapes, or regress the target distribution given input image conditions.
Our method achieves superior performance on both tasks over all the baselines.

\section{Related Work}

\paragraph{Normalizing flows on Euclidean space}

Most of the Normalizing flows are constructed in Euclidean space. Many of them are constructed using coupling layers\cite{dinh2014nice,dinh2016density}, and Glow\cite{kingma2018glow} improves it using $1 \times 1$ convolution for flexible permutation. Flow++\cite{ho2019flow++} combines multiple cumulative distribution functions to make the transformation in the coupling layer more expressive. Invertible ResNet\cite{behrmann2019invertible} and Residual Flow\cite{chen2019residual} propose residual flow which is more flexible and is also possible to be extended to $\SO$. Neural ODE\cite{chen2018neural} and FFJORD\cite{grathwohl2018ffjord} treat the transformation as a continuous movement of the vectors \haoran{with ordinary differential equation (ODE),} 
and RNODE\cite{finlay2020train} further improves it by adding constraints to \haoran{smooth the dynamics.} 

\noindent\textbf{Normalizing flows on non-Euclidean, SO(3)-relevant manifolds}
\haoran{ReLie\cite{falorsi2019reparameterizing} and ProHMR\cite{kolotouros2021probabilistic} both perform normalizing flow on the Euclidean space and then use a map from $\mathbb{R}^N$ to $\SO$, however, they suffer from either discontinuity or infinite-to-one mapping due to $\SO$'s special topological structure as shown in Supplementary Material.} 
Rezende et.al.\cite{rezende2020normalizing} propose three methods to construct flow on tori and sphere: Mobius transformation, circular splines, and non-compact projection, and here we use Mobius transformation \haoran{to build our flow on $\SO$.} 
Also, Mathieu et al.\cite{mathieu2020riemannian}, Lou et al.\cite{lou2020neural} and Falorsi et al.\cite{falorsi2020neural} extend continuous normalizing flow to Riemann manifold by parameterizing the ``velocity'' in the tangent space \haoran{and this can also be applied to $\SO$.} 
\haoran{Moser Flow\cite{rozen2021moser} further improves it by parametrizing the density as the prior density minus the divergence of a learned neural network, however, the model's effectiveness is limited when density is too high or too low, as it models probability instead of log probability. }

\noindent\textbf{Distributions for rotation} Several works leverage probabilistic distributions on $\SO$ for the purpose of rotation regression. 
Prokudin et al.\cite{prokudin2018deep} use the mixture of von Mises distributions over Euler angles. Gilitschenski et al.\cite{gilitschenski2019deep} and Deng et al.\cite{deng2022deep} utilize Bingham distribution over quaternion to jointly estimate a distribution over all axes.
Mohlin et al. \cite{mohlin2020probabilistic} leverage matrix Fisher distribution for deep rotation regression with unconstrained Euclidean parameters, which is also used in semi-supervised learning \cite{yin2022fishermatch}. A concurrent work \cite{yin2023laplace} proposes Rotation Laplace distribution inspired by multivariate Laplace distribution. 
Different from the parametric distributions above, Murphy et al. \cite{murphy2021implicit} 
represents distributions implicitly by neural networks, where it predicts unnormalized log probability first and  then normalize it by discretization sampling on $\SO$.
In this work, we use normalizing flows to directly generate normalized distributions.

\section{Normalizing Flows on Riemannian Manifold}

Normalizing flows (NFs) provide a flexible way to construct complex distributions in high-dimensional Euclidean space by transforming a base distribution through an invertible and differential mapping. \yulin{Base distributions are often chosen to be easily evaluated and sampled from, like gaussian distribution in Euclidean space.} \yulin{NFs} can be extended to Riemannian manifolds embedded in a higher dimensional space \cite{papamakarios2021normalizing,gemici2016normalizing}. Formally, normalizing flows transform base distributions $\pi(u), u\in \mathcal{M}$ to target distributions $p(x), x\in \mathcal{N}$, where $\mathcal{M,N}$ are Riemannian manifold and have the same topology, via diffeomorphisms $T: \mathcal{M}\rightarrow\mathcal{N}$. The probability density function(pdf) of $x$ can be calculated by change of variable formulas:
\begin{equation}
    p(x)=\pi(T^{-1}(x))|\det J_{T^{-1}}(x)|\label{eq:change of variable},
\end{equation}
where Jacobian matrix $J_{T^{-1}}(x)=\frac{\partial(T^{-1}(x))}{\partial x}$ is the $D \times D$ partial derivatives of $T^{-1}$ at $x$.
\par
As diffeomorphisms are composable, in practice, the transformation $T$ is often implemented via a sequence of simple transformations $T = T_K\circ\cdots \circ T_2\circ T_1$, whose Jacobian determinants are easy to evaluate. The  determinant of the composed transformation is given by:
\begin{equation}
    \Det{J_{T^{-1}}(x)}=\Pi_{i=1}^{k} \Det{J_{T_i^{-1}}(T_K^{-1}\circ\cdots\circ T_{i+1}^{-1}(x))}
\end{equation}

Normalizing flows enable both forward and inverse processes and one can calculate $p(x)$ through the process. We can fit the target distribution by minimizing the negative log-likelihood (NLL) of training data in the inverse process and sample via mapping samples from the base distribution in the forward process.

\par

\section{Method}
Our normalizing flows comprise two key building modules that operate on the $\SO$ manifold. First, we utilize a Mobius coupling layer that interprets the rotation matrix as three orthonormal 3-vectors, with one vector held fixed \haoran{and the remaining part transformed using the Mobius transformation with the fixed vector as condition. } 

Second, we employ a quaternion affine transformation that uses the quaternion representation while retaining antipodal symmetry, making it a diffeomorphism on real projective space $\mathbb{RP}^3$. Affine transformation resembles the $1\times1$ convolution in Glow\cite{kingma2018glow}, but is applied to the quaternion representations rather than the channels. 

\begin{figure*}
    \centering
    \includegraphics[width=0.8\linewidth, angle=0]{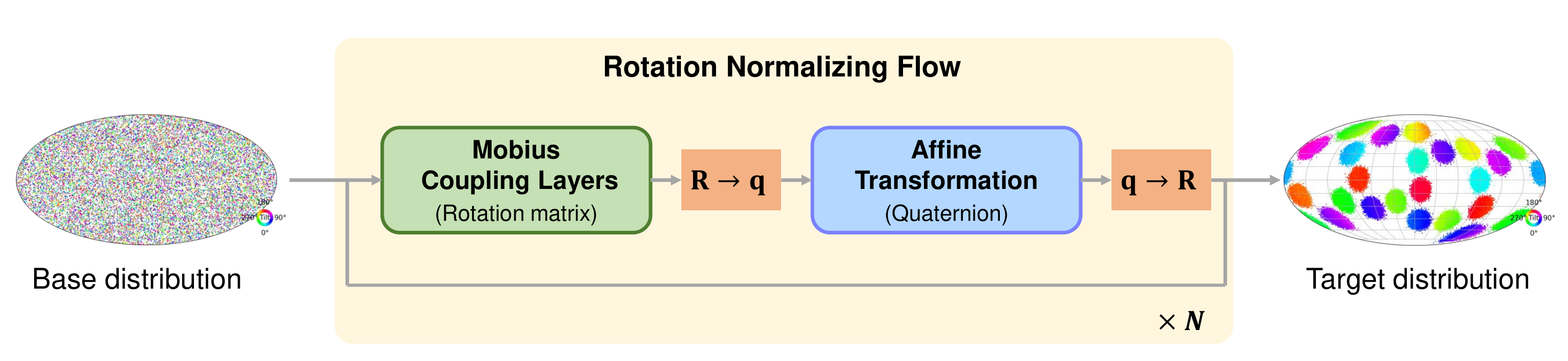}
    \caption{\textbf{Pipeline overview.} Our flow model takes rotations as input and outputs transformed rotations and log determinants of Jacobian, transforming a \yulin{base} distribution to a target one. 
    Our flow is done by iteratively alternating Mobius coupling on Rotation matrix representation and affine transformation on quaternion for N times. For probability inference, data are fed into the flow to the corresponding rotation which is of \yulin{base} distribution and predicts log-likelihood; while in the sampling process, our flow runs inversely, generating new data by transforming samples from the \yulin{base} distribution. The distribution visualization is borrowed from \cite{murphy2021implicit} where $\SO$ is projected to a 2D sphere by Hopf fibration, points on the 2D sphere indicate the direction of a canonical z-axis, the colors represent the tilt angle about that axis, the direction of a canonical z-axis and the sizes of points show the probability density.}
    \label{fig:pipeline}
\end{figure*}

Mobius coupling layers are generally effective at modeling various distributions, while affine transformations are more suitable for uni-modal distributions. By combining these two building blocks, we can construct more powerful normalizing flows that perform better and converge faster. Our model consists of multiple blocks of layers, each of which comprises a Mobius coupling layer and a quaternion affine transformation. Figure \ref{fig:pipeline} illustrates our model.

\par

\subsection{Mobius Coupling Layer}
Mobius Transformation is defined on a D-dimensional sphere $\mathcal{S}^D$. Rezende et al.\cite{rezende2020normalizing} have applied it to build expressive normalizing flows on the 2D circle $\mathcal{S}^1$. However, with the unique topology of $\SO$, it's non-trivial to apply Mobius transformation to $\SO$ manifold. We present a coupling layer method that fully utilizes the geometry of $\SO$ and provides a $\frac{\sqrt{2}}{2}$ trick to solve the discontinuity encountered in combining multiple transformations. 
\par
\noindent\textbf{Revisit Mobius transformation} Mobius Transformation on $\mathcal{S}^D$ can be parameterized by an $\omega\in \mathbb{R}^{D+1}$ \haoran{satisfying $\Vert\omega\Vert < 1$.} 
For a point $c \in \mathcal{S}^D$, Mobius transformation $f_\omega$ is defined as:
    \begin{equation}
        f_\omega(c)=\frac{1-\Vert \omega\Vert^2}{\Vert c- \omega\Vert}(c-\omega)-\omega
    \end{equation}

\begin{figure}
    \centering
    \vspace{1pt}
    \includegraphics[height=25mm]{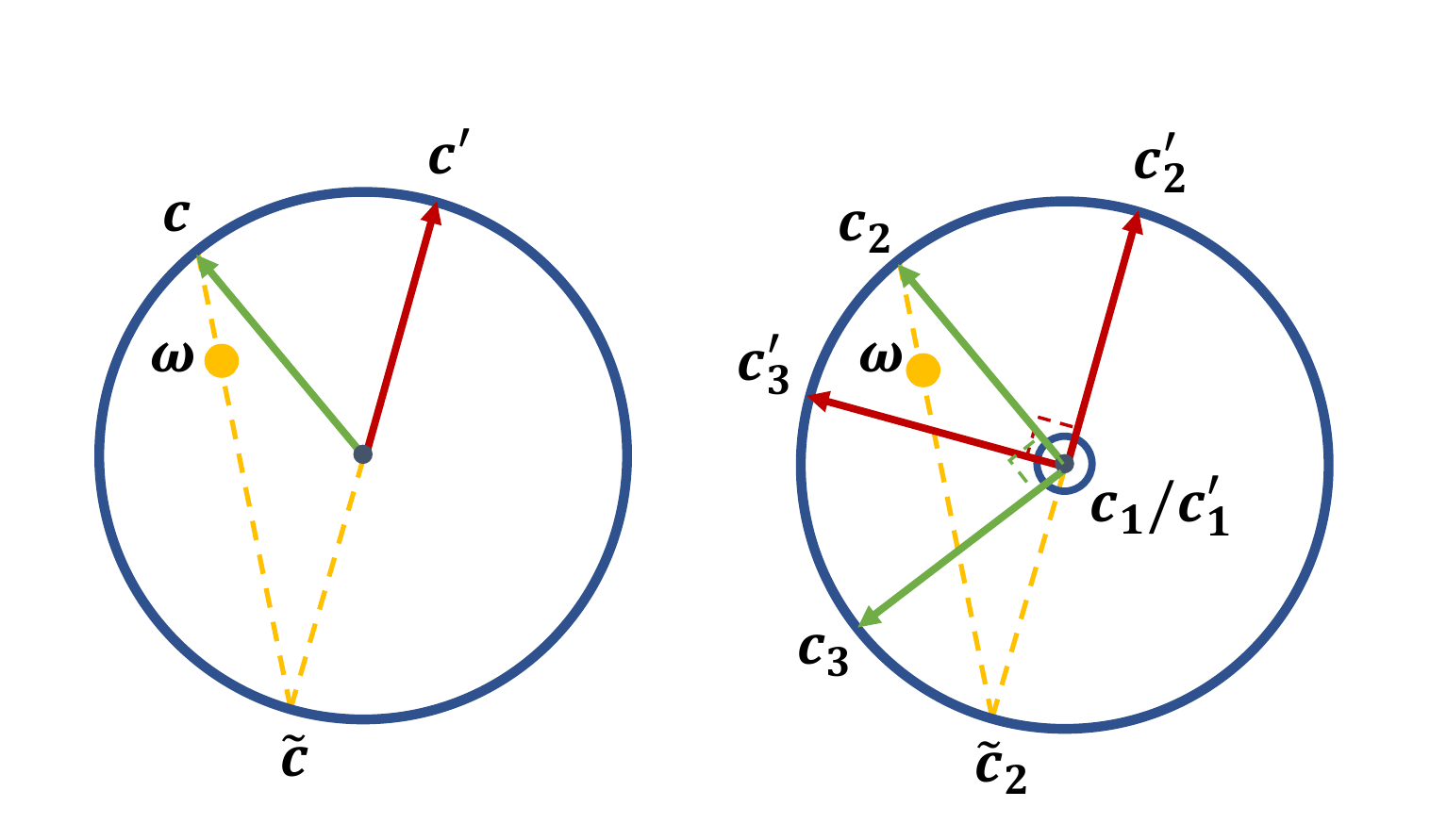}
    \hspace{8mm}
    \includegraphics[height=20mm]{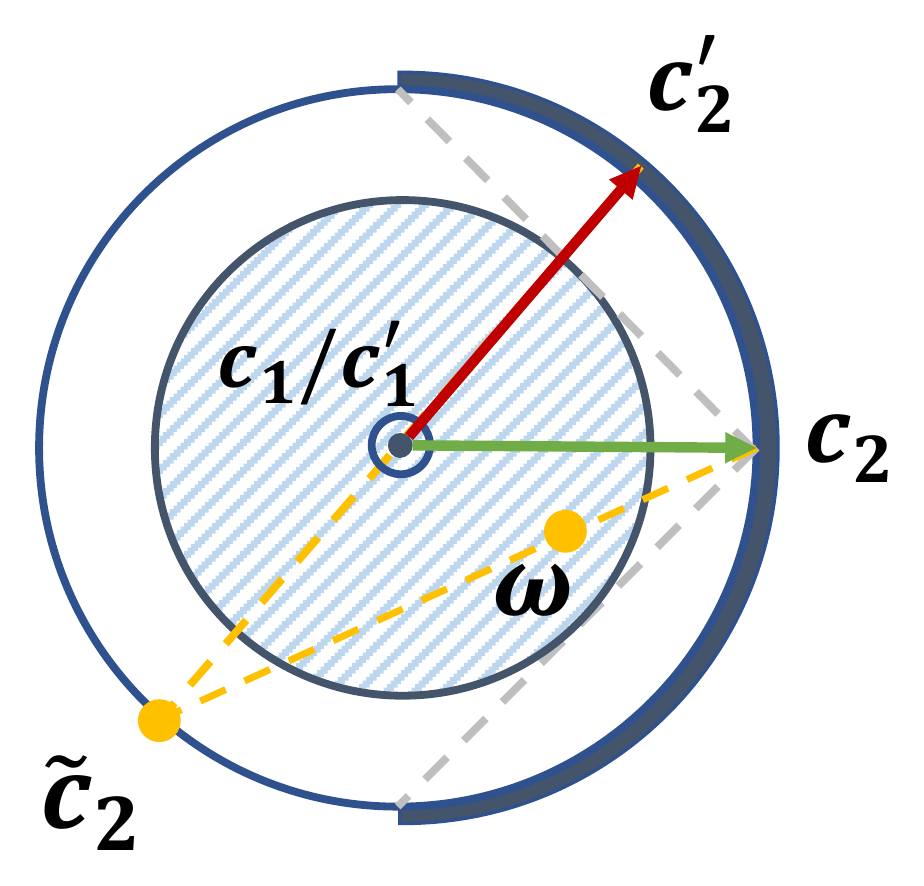}
    \caption{\textbf{Left}: Geometric illustration of Mobius \haoran{transformation.} The transformation is implemented by first connecting a straight line between $c$ and $\omega$ which intersects the sphere with $\tilde{c}$\haoran{, and the result} $c^\prime$ is the opposite point of $\tilde{c}$. \haoran{\textbf{Middle}: Geometric illustration of Mobius coupling layer.} This sketch is viewed parallel to the unchanged $c_1$. An MLP conditioned on $c_1$ gives output $\omega^\prime$ which is then projected to $\omega$ orthogonal to $c_1$. $c_2$ is then transformed \haoran{into $c_2^\prime$} via Mobius transformation with parameter $\omega$. $c_3^\prime$ is then computed by cross product of \haoran{$c_1^\prime=c_1$} and $c_2^\prime$. 
    \textbf{Right}: Illustration of $\frac{\sqrt{2}}{2}$ trick. $\omega$ are restricted in the blue circle and resulting in $\{{c_2^\prime}_i\}$ on the semi-circle of blue lines. Therefore the combined $c_2^\prime$ are restricted on the semi-circle of blue lines. Thus slight change of $\{\theta_i\}$ won't cause jump (about $\pi$) in $\theta$, solving the discontinuity.
    \vspace{-4mm}}
    \label{fig:trick}
\end{figure}

This transformation has a very straightforward geometric meaning: \haoran{first extend the line $c\omega$ and find the intersection point between it and the sphere $\tilde{c}$, then output the point symmetric to $\tilde{c}$ about the origin $c'=-\tilde{c}$, as shown in Figure \ref{fig:trick} Left.} 
When $\omega$ is at the origin, $f_\omega$ becomes identity transformation; when $\omega$ is not at the origin, $f_\omega$ concentrates part away from $\omega$; and when $\omega$ is near to surface of unit sphere $\mathcal{S}^D$, $f_\omega$ maps almost all points on $\mathcal{S}^D$ to nearing of $-\omega$.

\begin{figure}
    \centering
    \vspace{1pt}
    \includegraphics[height=30mm]{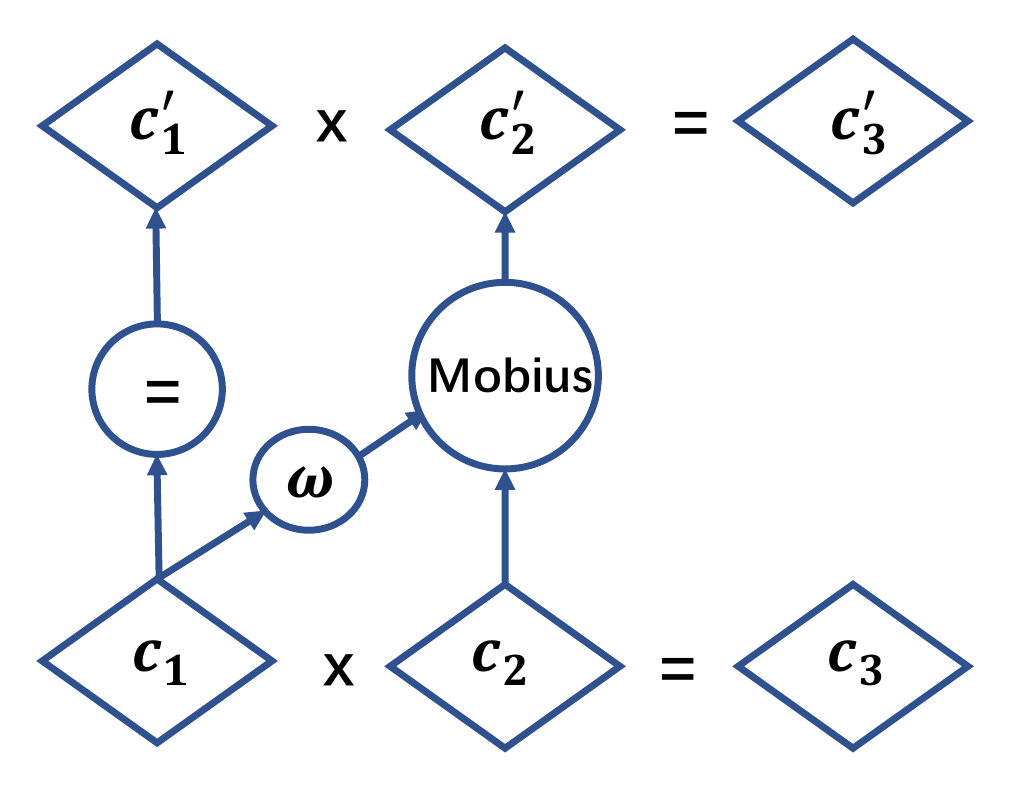}\hspace{0.05\columnwidth}\includegraphics[height=30mm]{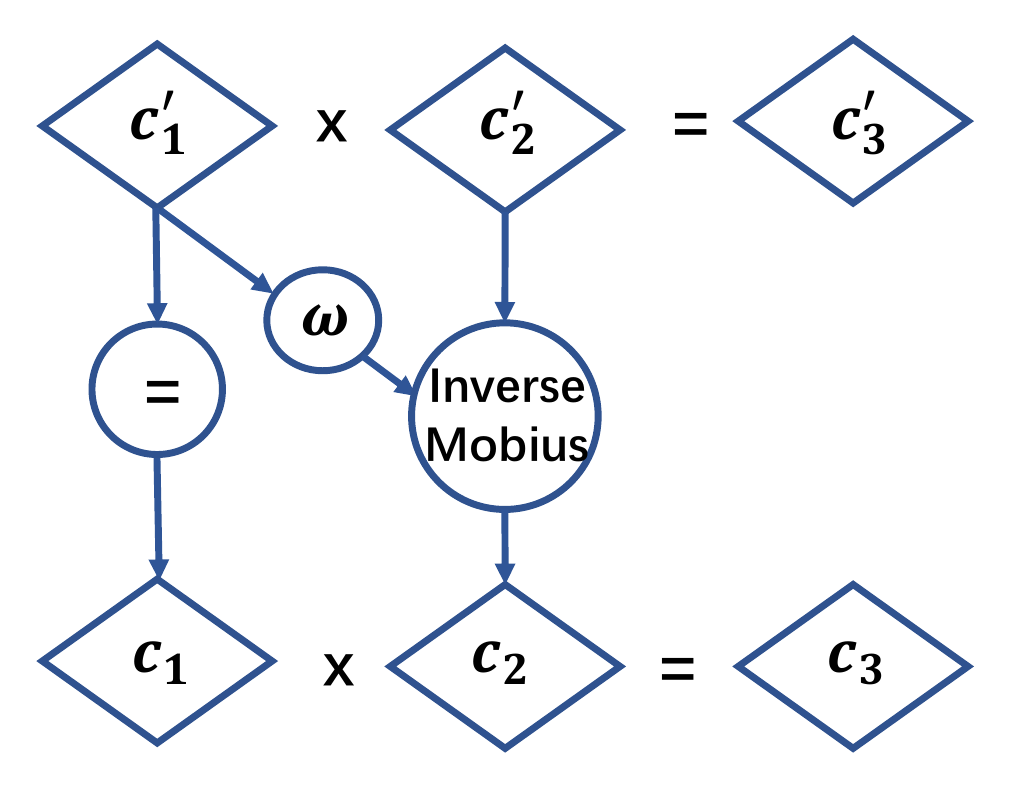}
    \caption{\textbf{Computational graphs of Mobius coupling layer.} $c_1, c_2, c_3$ indicate different columns for rotation matrix and $c_1^\prime, c_2^\prime, c_3^\prime$ indicate the columns for transformed rotation matrix. Mobius coupling layer applies the Mobius transformation to column $c_2$ conditioned on the unchanged column $c_1$. 
    \vspace{-4mm}}
    \label{fig: Mobiusillu}
\end{figure}

\noindent\textbf{Mobius coupling layer} 
To build a normalizing flow on $\SO$, we thus consider applying the idea of \textit{coupling} layer introduced in \cite{dinh2014nice,dinh2016density} to the orthonormal vectors. In each \textit{coupling} layer, the input $x$ is divided into the \textit{conditioner} $x_1$ that remains unchanged after the flow and the \textit{transformer} $x_2$ that changes according to the condition part, which can be written as $x^\prime_1=x_1, x_2^\prime=f(g(x_1), x_2)$. As the calculation of the Jacobian determinant only involves $\frac{\partial f}{\partial x_2}$, $g$ can be arbitrarily complex which enables high expressivity \haoran{and here we use a neural network to parameterize it}.

A $3\times3$ rotation matrix $R \in \SO$ satisfies $RR^T = I$ and $\det{R} = +1$. 
It thus can be expressed as three orthonormal vectors $[c_1, c_2, c_3]$ that satisfy $||c_i|| = 1$ (\haoran{meaning }
$c_i \in \mathcal{S}^2$) and $c_i \cdot c_j = 0$ for all $i \neq j$.

We utilize a similar structure to build Mobius transformation for rotation matrices. We divide a rotation matrix into 2 parts, the \textit{conditioner} is $c_i$ ($i=1,2,3$) and the \textit{transformer} is the rest two columns $\{c_j~|~j\neq i\}$. 
Taking $i=1$ as an example: conditioning on $c_1$, we can transform $c_2$ to $c_2'$. Then $c_3'$ is already determined by $c_3' = c_1' \times c_2'$\haoran{ where $c_1'=c_1$}. The coupling layer needs to ensure that: 1) $||c_2'|| = 1$, \textit{i.e.} $c_2'\in \mathcal{S}^2$; and 2) $c_2'$ is orthogonal to $c_1'$.

Given condition 1), we thus consider using a Mobius transformation on $\mathcal{S}^1$ to transform $c_2$.
To further meet condition 2), we notice that all valid $c_2$ and $c_3$ form a plane $P$ that passes the origin \haoran{and is perpendicular to $c_1$.} 
After the transformation, $c_2'$ needs to stay in $P$. 
This can be achieved by constraining $\omega$ inside $P$.
Therefore, we propose to learn a neural network that maps the condition $c_1$ to $\omega'\in\mathbb{R}^3$ and then projects it to $\omega\in P$, as shown below: 
\begin{equation}
    \omega = \omega' - c_1 (c_1\cdot \omega')
\end{equation}
where $\omega'$ is the unconstrained parameters generated by the neural network. The structure of our Mobius coupling layer is illustrated in Figure \ref{fig: Mobiusillu} \haoran{and Figure \ref{fig:trick} Middle}.
\par
Note that, given $c_1$, there is only 1 Degree of Freedom left for the rest two columns. So, our Mobius coupling layer is essentially rotating $c_2$ and $c_3$ about $c_1$ simultaneously by an angle $\theta $ conditioned on $c_1$\haoran{ and $c_2$}. 

\noindent\textbf{Linear combination of multiple Mobius transformations} 
To further increase the expressivity of the Mobius transformation, we leverage linear combination of \haoran{Mobius} transformation presented in \cite{rezende2020normalizing}. \haoran{It is done by first transforming $c_2$ into $\{{c_2^\prime}_i\}$ using a set of $\omega$s, $\{\omega_i\}$, and then calculate the weighted sum $\theta^\prime=\sum \alpha_i \theta_i$, where $\alpha_i$ is the weight generated by a neural network conditioned on $c_1$ and normalized with softmax, and $\theta_i$ is the angle between ${c_2^\prime}_i$ and $c_2$. The result of combined transformation $c_2^\prime$ can then be calculated by rotating $c_2$ with $\theta^\prime$. }
\par
However, such naive implementation has the problem of discontinuity. Take two combination points with weights [0.5, 0.5] for example. Assume $\theta_1$ is 0$^{\circ}$, $\theta_2$ is $-178^{\circ}$, the combined angle $\theta$ is $-89^{\circ}$. However, when $\theta_2$ slightly changed to $-182^{\circ}$ that is $178^{\circ}$ as $\theta\in[-\pi, \pi]$, the combined angle $\theta$ becomes $89^{\circ}$. This discontinuity of slight change of $\theta_i$ resulting in a huge jump in combined $\theta$ can reduce the networks' performance and add difficulties in learning. 
\par
\noindent\textbf{$\frac{\sqrt{2}}{2}$ trick}
\haoran{We present a $\frac {\sqrt{2}}{2}$ trick to alleviate this discontinuity. If $\Vert\omega\Vert$ is small, $c_2^\prime$ will be close to $c_2$. And if all $\{{c_2^\prime}_i\}$ lie on a small neighborhood of $c_2$, there won't be discontinuity issues as circle is locally linear, whereas the expressiveness will be limited if $\{{c_2^\prime}_i\}$ are too close to $c_2$. It can be shown that the biggest range of $c_2^\prime$ with no discontinuity in combination is a semi-circle whose center is $c_2$ and the corresponding range of $\Vert\omega\Vert$ is $[0, \frac{\sqrt{2}}{2})$ as shown in Figure \ref{fig:trick} Right.}

\par
By linear combination, the inverse of transformation can't be calculated analytically, we alleviate binary search presented in \cite{rezende2020normalizing} to find the inverse as $\theta \in(-\pi/2, \pi/2)$. Though the restriction of $\Vert\omega\Vert$ may reduce the expressivity of our flows, the avoidance of discontinuity stabilizes our network so in general it is beneficial. (See Supplementary Material for details)

\subsection{Quaternion Affine Transformation}\label{sec: Affine Transformation}
\noindent\textbf{Bijective transformation on $\SO$ manifold}
Quaternion is another representation of rotation, defined as a unit vector on the 4D sphere. Let $\theta$ be the angle of the rotation and $(x, y, z)$ be the axis of rotation, then $\mathbf{q}$ can be computed as $(\cos\frac{\theta}{2}, x\sin\frac{\theta}{2}, y\sin\frac{\theta}{2}, z\sin\frac{\theta}{2})$.
\par

\haoran{However, quaternion representation of $\SO$ has} the topology of antipodal symmetry, \haoran{meaning} that $\mathbf{q}$ and  $-\mathbf{q}$ represent the same rotation $R$. To be bijective on $\SO$, transformation on quaternion should keep antipodal symmetry, i.e. transform $-\mathbf{q}$ to $-f(\mathbf{q})$. Our proposed affine transformation needs to satisfy such requirement in order to be a bijective transformation on $\SO$. 
\par
\noindent\textbf{Affine transformation}
Our proposed quaternion affine transformation consists of a linear transformation of 4D vectors $\mathbf{q}$ followed by a projection to the unit sphere $\mathcal{S}^3$. The explicit expression for this transformation is as follows:
\begin{equation}
    g(\mathbf{q})=\frac{W\mathbf{q}}{\Vert W\mathbf{q}\Vert}\label{eq:affine}
\end{equation}
\begin{figure}
    \centering
    \vspace{1pt}
    \includegraphics[width=\columnwidth]{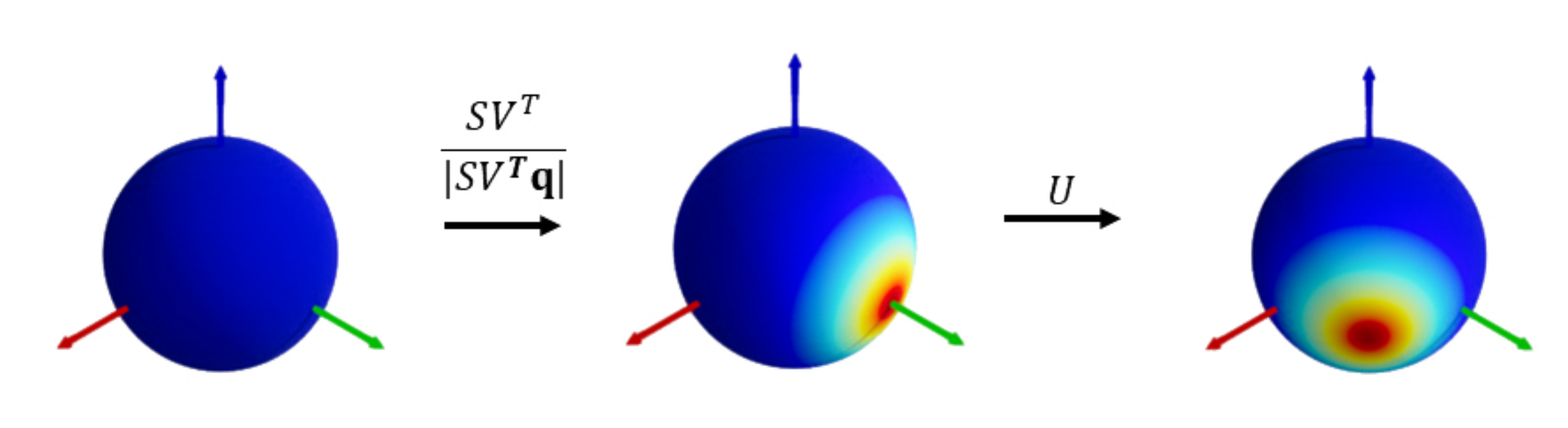}
    \includegraphics[height=18mm]{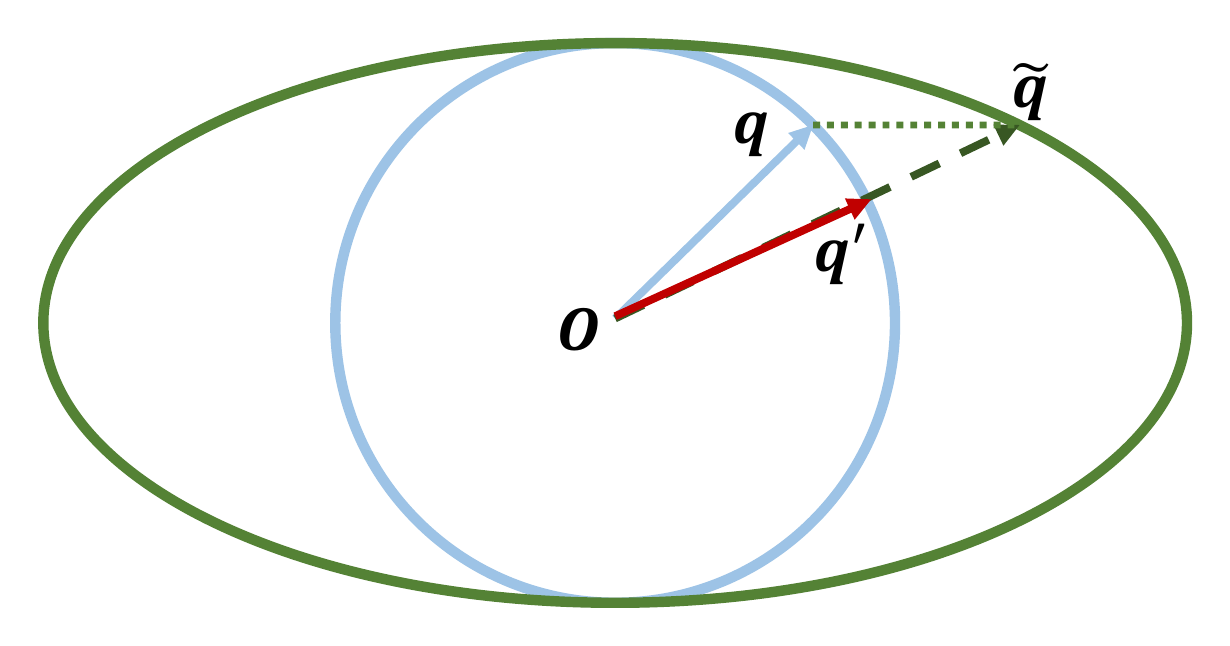}
    \hspace{2mm}
    \includegraphics[height=18mm]{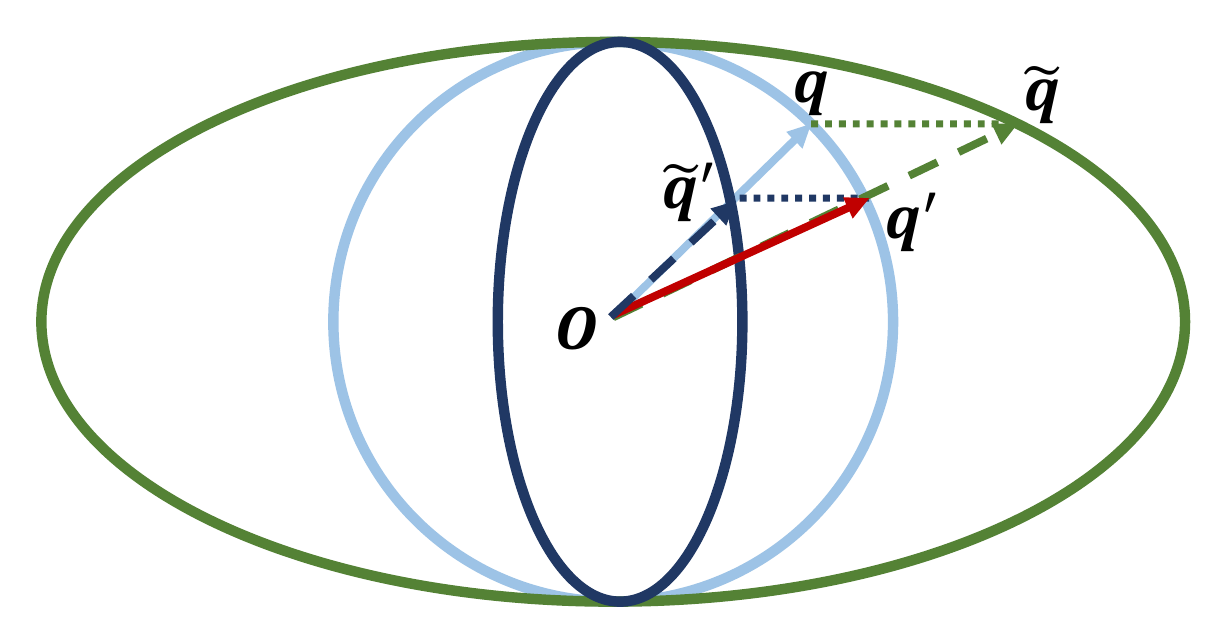}
    \caption{\textbf{Illustration of quaternion affine transformation.} \textbf{Top}: Effect of quaternion affine transformation. Let $e_1, e_2, e_3, e_4$ correspond to the standard basis of $\mathcal{S}^3$, i.e. $w, x, y, z$ unit vectors. The affine transformation first rotates $\mathbf{q}$ by $V^T$, followed by the scaling and normalizing part which concentrates points near the axis $V^T e_i$ with large $s_i$ and expands the others, and followed by a rotation $U$. \textbf{Bottom}: Forward and inverse sketch of scaling and normalization part in affine transformation. The sketch is viewed in 2D $w-x$ section. $\mathbf{q}$ is transformed to $\tilde{\mathbf{q}}$ by multiplying scaling factors $s_1=2, s_2=1, s_3=1, s_4=1$, and then normalized to $\mathbf{q}'$. The inverse transformation is similar to forward process with $s_1^\prime=1/s_1, s_2^\prime=1/s_2, s_3^\prime=1/s_3, s_4^\prime=1/s_4$. This feature is closely related to the geometry of affine transformation of an ellipse and more discussions are given in Supplementary Material.}
    \label{fig:affine}
    \vspace{1pt}
\end{figure}
where $W$ is a $4\times 4$ invertible matrix, $\mathbf{q}$ is the quaternion representation of a rotation. The inverse of the transformation \haoran{can be calculated}
by \haoran{simply} replacing $W$ by its inverse $W^{-1}$. The transformation looks similar to the $1\times 1$ convolution in Glow\cite{kingma2018glow}, in which an invertible matrix is multiplied on picture channels. We present a geometric explanation for its effect and explain the meaning of its name. 
\par
\noindent\textbf{Geometric explanation of affine transformation}
We name this transformation  \textit{affine}, since it resembles 
the affine transformation in Eucliean space $f = ax + b$,
where $a$ is the scaling parameter, and $b$ is a displacement term. By SVD decomposition $W=USV^T$, the $4 \times 4$ invertible matrix can be decomposed into an orthogonal matrix $U$, multiplied by a diagonal matrix $S$ and another orthogonal matrix $V^T$. Multiplying $U,V^T$ globally rotates the distribution pattern and acts as a displacement on $\SO$ manifold, while the diagonal matrix $S$ serves as the scaling term. 
\par
As multiplying an orthogonal matrix will not change the length of a vector, the term $\Vert W\mathbf{q} \Vert$ is equal to $\Vert S(V^T\mathbf{q}) \Vert$, so affine transformation can be decomposed into:
\begin{equation}
    g(\mathbf{q})=U\frac{S}{\Vert S(V^T\mathbf{q}) \Vert}(V^T\mathbf{q})
\end{equation}
This can be seen as a 4-step transformation, first \haoran{rotate $\mathbf{q}$ to $V^T\mathbf{q}$, then multiply each coordinate by scaling factors $s_{i}, (i=1, \cdots, 4)$ and normalize it to a unit vector}, and finally rotate the quaternion with $U$. An illustration of quaternion affine transformation is shown in Figure \ref{fig:affine} Top.

\par
We present 2 methods to parameterize $4\times4$ invertible matrix $W$. The first one is to simply use an unconstrained $4\times4$ matrix as 
it is very rare for the matrix to be (or near) singular, and we find it stable in our experiments.
We also tried LU decomposition $W=PL(U+S)$ as presented in \cite{kingma2018glow}, where $P$ is a fixed permutation matrix, $L$ is a lower triangular matrix with ones on the diagonal, $U$ is an upper triangular matrix with zeros on the diagonal, and $
S$ is a vector with non-zero coordinate. See more discussions in Supplementary Material. 
\par
\noindent\textbf{Why rotation $U, V$ is needed?}  
Our Mobius coupling layer allows distribution to flow in the vertical plane of $c_i$, however, it is very difficult to learn a global rotation of distributions on $\SO$. The introduced rotation operation in quaternion space exactly alleviates this problem. 
Rotating quaternions also serve as a generalization of permutation in splitting condition or transformed columns of rotation matrix, which has similar effects to the $1\times 1$ Convolution introduced in Glow \cite{kingma2018glow}.

\noindent\textbf{Why scaling and normalization?}  
Multiplying diagonal matrix $S$ results in multiplying coordinate of a quaternion $\mathbf{q}=(w, x, y, z)$ by scaling $s_1, s_2, s_3, s_4$,
\begin{equation}
    (w, x, y, z) \xrightarrow{} (s_1 w, s_2 x, s_3 y, s_4z)
\end{equation}
transforming the unit sphere to an ellipsoid, as shown in Figure \ref{fig:affine} Bottom. A point $\mathbf{q}$ on the sphere $\mathcal{S}^3$ is transformed to a point on the oval $\mathbf{\tilde{q}}$. It is followed by a projection to sphere: term $1/\Vert S\mathbf{\tilde{q}}\Vert $ normalizes the transformed vector. The final point $\mathbf{q}^\prime$ is the intersected point of $O-\mathbf{\tilde{q}}$ on the sphere $\mathcal{S}^3$.
\par
The explicit expression for scaling and normalization is:
\begin{equation}
    f_{(s_1,s_2,s_3,s_4)}(w, x, y, z)=\frac{(s_1w, s_2x, s_3y, s_4z)}{\vert\vert (s_1w, s_2 x, s_3 y, s_4 z )\vert\vert}
\end{equation}
\par

When $s_1=s_2=s_3=s_4$, the scaling and normalization transformation is identity\haoran{, otherwise it concentrates probability to the axis with large $s$.} 
Scaling and normalization can create a high peak by transforming $\mathcal{S}^3$ to an elongated ellipsoid and then projecting back to $\mathcal{S}^3$. 
\par

\subsection{Interchange of Rotation Matrix and Quaternion Representation}
We iteratively compose Mobius coupling layer and quaternion affine transformation to build our flow model \haoran{and switch between rotation matrix representation and quaternion representation in the process}. 
\par
The composed flows perform better and learn faster, as the quaternion affine transformation makes up for some problems in Mobius coupling layer, and increases its expressiveness. The rotation operation in quaternion affine transformation allows a global rotation of distributions on $\SO$, and serves as a generalization of permutation in splitting condition or transformed columns of rotation matrix. The scaling and normalization operation makes it possible to quickly create high peak distributions. This allows for quick concentrate distribution to target predictions and accelerates the convergence concerning rotation regression.

\subsection{Conditional normalizing flow}
\label{sec:conditional-nf}
There are cases when we need to infer a distribution depending on condition, for example when we need to infer the rotation of a symmetric or occluded object in an image.
Our flow can be easily extended to be conditional using methods in \cite{winkler2019learning} as we can simply concatenate the condition with fixed columns to generate parameters for Mobius flow\haoran{, and use a neural network to output the matrix $W$ in quaternion affine transformation. Moreover, apart from using uniform distribution as base distribution, we can pretrain a neural network to fit the target distribution using commonly-used unimodal distributions (e.g. Matrix Fisher distribution \cite{mohlin2020probabilistic}) as the base distribution. } 
We conduct experiments with our conditional normalizing flows on learning multimodal rotation distributions for symmetric object \yulin{and also common unimodal tasks.} We extract features by ResNet given input images. 
\section{Experiments}

In this section, we conduct multiple experiments to validate the capacity of our proposed normalizing flows to model distributions on $\SO$. We train all experiments with negative log-likelihood (NLL) loss. The invertible matrix $W$ for quaternion affine transformation is parameterized as a $4\times 4$ unconstrained matrix. Implementation details and results of LU decomposition parameterization for $W$ are reported in Supplementary Material.

\subsection{Learning to Model Various Distributions}
\label{sec:unconditional}
As in common, we first evaluate and compare our normalizing flows and baseline methods by learning to fit distributions on $\SO$ with distinct properties. 

\noindent\textbf{Datasets}  We design four challenging distributions on $\SO$: a very sharp single modal distribution, a 24-peaked multi-modal distribution, a cyclic distribution, and a 3-line distribution. 
The 24-peaked distribution and the cyclic distribution are designed to simulate the symmetry property of \textit{cube} and \textit{cone} solids. We adopt the visualization tool of \cite{murphy2021implicit} and show the target distributions as follows.

\noindent\textbf{Baselines} 
\cite{falorsi2019reparameterizing} introduces the reparameterization trick for Lie groups and allows for constructing flows on the Lie algebra of $\SO$. \cite{mathieu2020riemannian} proposes continuous normalizing flows on Riemannian manifold, and we apply it to $\SO$ manifold. \cite{murphy2021implicit} models the distribution implicit by the neural networks, where the $\SO$ space is uniformly discretized. Finally, we compare the mixture of matrix Fisher distribution with 500 components.

\noindent\textbf{Results}  
The results are reported in Table \ref{tab:unconditional}, where our model(Mobius + Affine) consistently achieves state-of-the-art performance among all baselines, demonstrating the ability of our method to fit arbitrary distributions in various shapes. We also report results of flows composed of single transformation, i.e. Mobius coupling layer or quaternion affine transformation. Results demonstrate that Mobius coupling layers can generally perform well while quaternion affine transformations are more suited for uni-modal distributions and infinite-modal distributions. Improvement by composing Mobius coupling layers and quaternion affine transformation is more demonstrated in conditional tasks, as shown in Sec. \ref{sec:ablation}.
\par
In our experiment, baseline \cite{falorsi2019reparameterizing} fails to fit the sharp distribution due to numerical unstable. For \haoran{detailed} 
discussions, see Supplementary Material. 

\setlength{\tabcolsep}{6.5pt}
\setlength{\tabcolsep}{6.5pt}
\begin{table}[!h]
\centering
\footnotesize
\caption{
\textbf{Comparisons on learning to fit various distributions.}  We adopt log-likelihood as the evaluation metric and use uniform distribution in $\SO$ as base distribution.
}
\resizebox{\columnwidth}{!}{
\begin{tabular}
{lccccc}
  \toprule
   log likelihood $\uparrow$ & {avg.}  & {peak} &{cone}     & {cube}    & {line}  \\  
    \midrule
    Riemannian\cite{mathieu2020riemannian} & 5.82 & 13.47 & 8.82 & 1.02 & -0.026 \\
    ReLie\cite{falorsi2019reparameterizing} & - & - & 5.32 & 3.27 & -6.97\\
    IPDF\cite{murphy2021implicit} & 4.38 & 7.30 & 4.75 & 4.33 &  1.12\\
    Mixture MF \cite{mohlin2020probabilistic}    & 6.04& 10.52 & 8.36 &  4.52 & 0.77 \\
    Moser Flow\cite{rozen2021moser} & 6.28 & 11.15 & 8.22 & 4.42 & \textbf{1.38}\\
    \hline
    Ours(Mobius) & \textbf{7.28}& \textbf{13.93}& 8.98 & \textbf{4.81} & \textbf{1.38}\\
    Ours(Affine) & 5.59& 13.50 & 8.84 & 0.00 & 0.00 \\
    Ours(Mobius+Affine)
    & \textbf{7.28} & \textbf{13.93} &  \textbf{8.99} &  \textbf{4.81}&  \textbf{1.38} 
    \\
  \bottomrule
\end{tabular}
}
\label{tab:unconditional}
\end{table}

\begin{figure}[htb]
    \centering
    \includegraphics[width=0.23\linewidth]{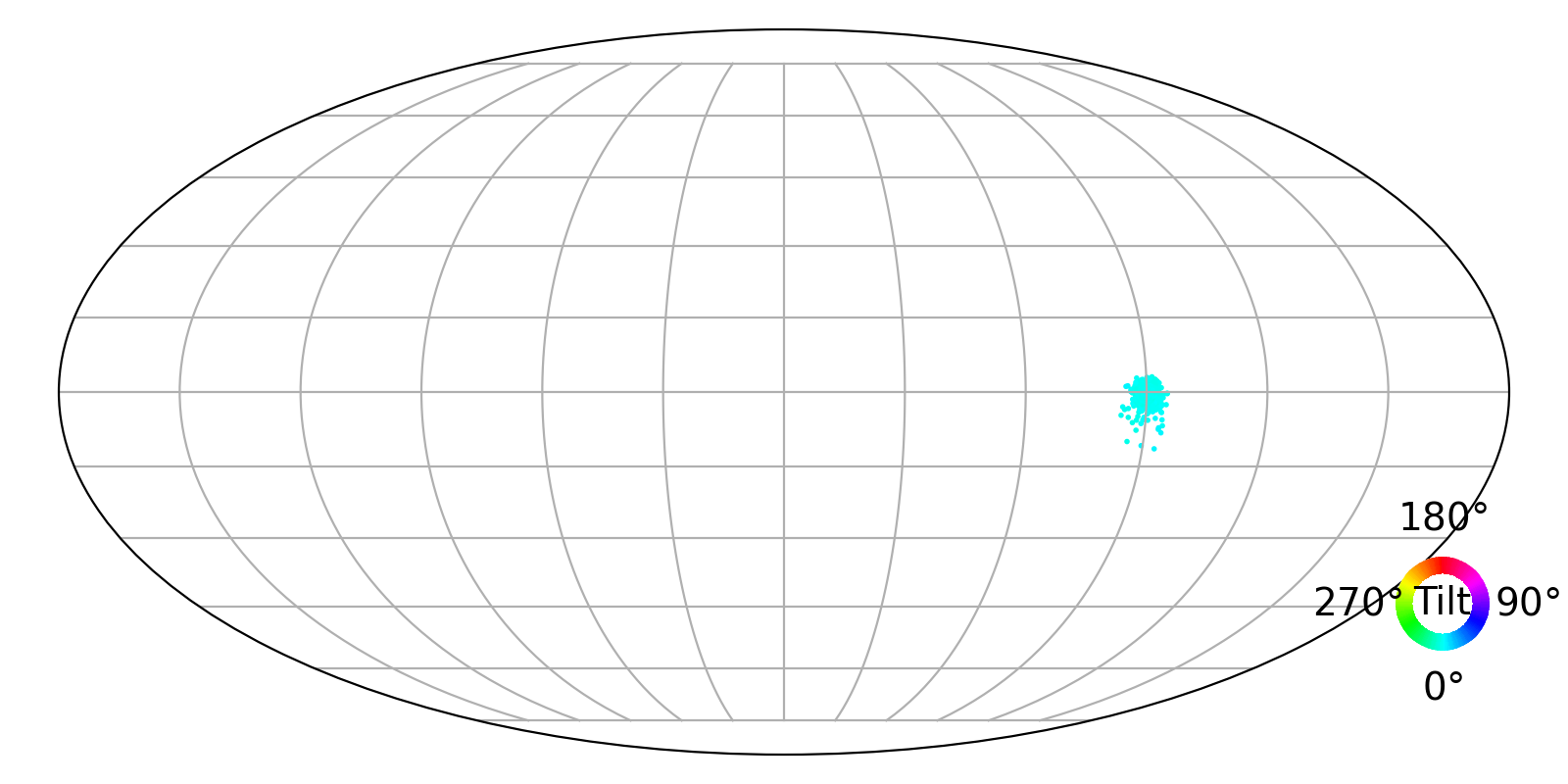} \hspace{0.5mm}
    \includegraphics[width=0.23\linewidth]{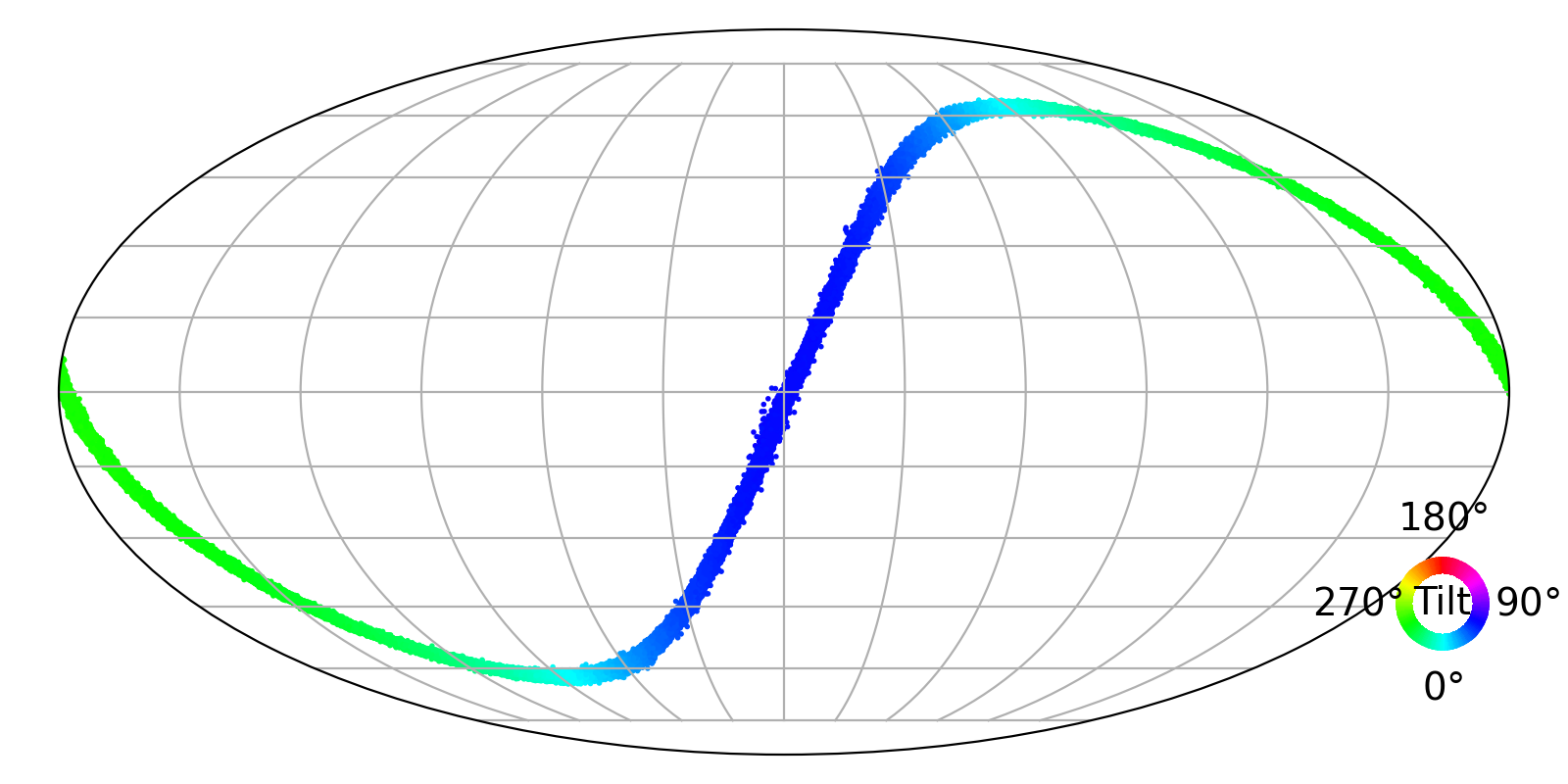} \hspace{0.5mm}
    \includegraphics[width=0.23\linewidth] {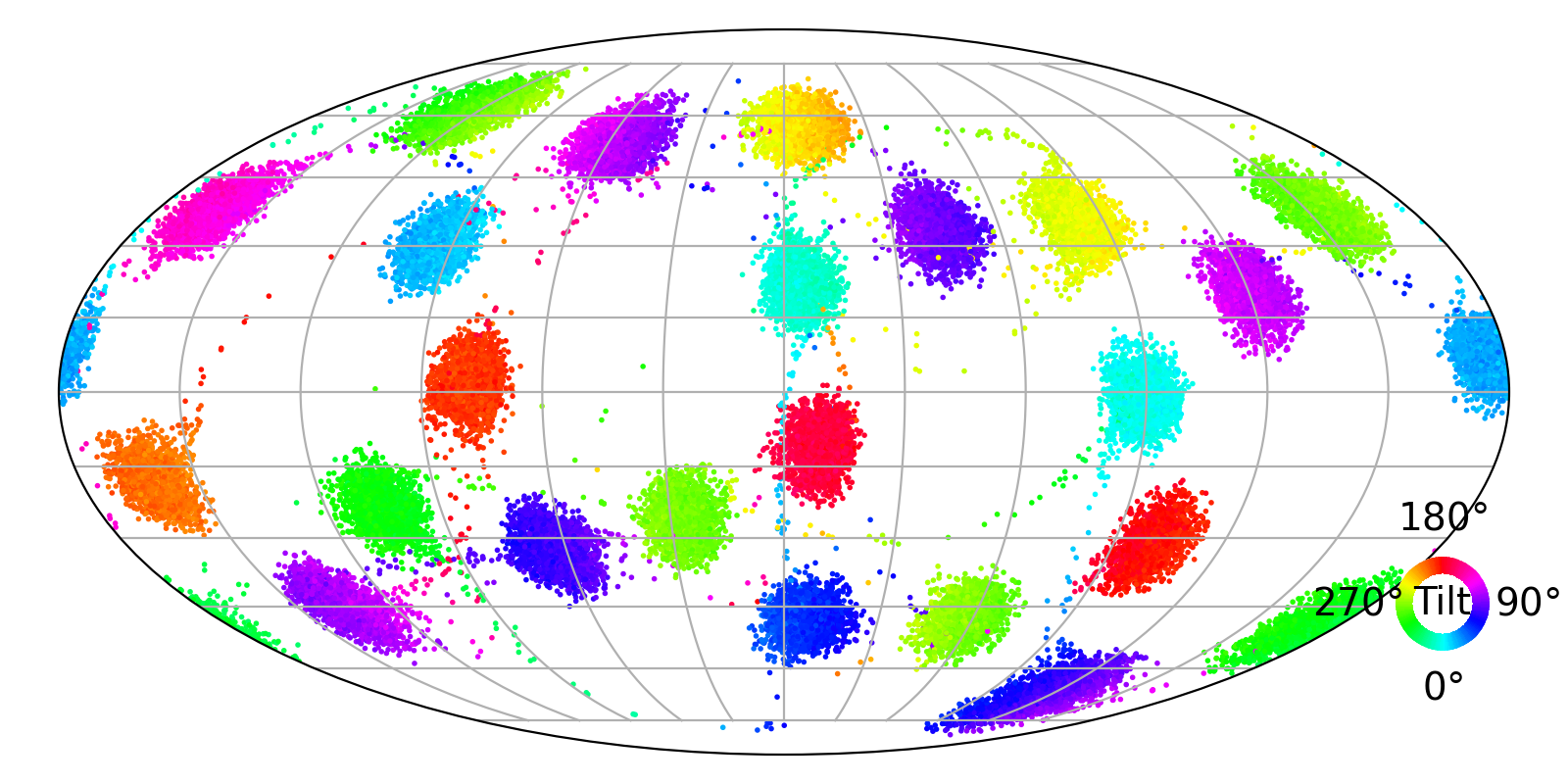} \hspace{0.5mm}
    \includegraphics[width=0.23\linewidth]{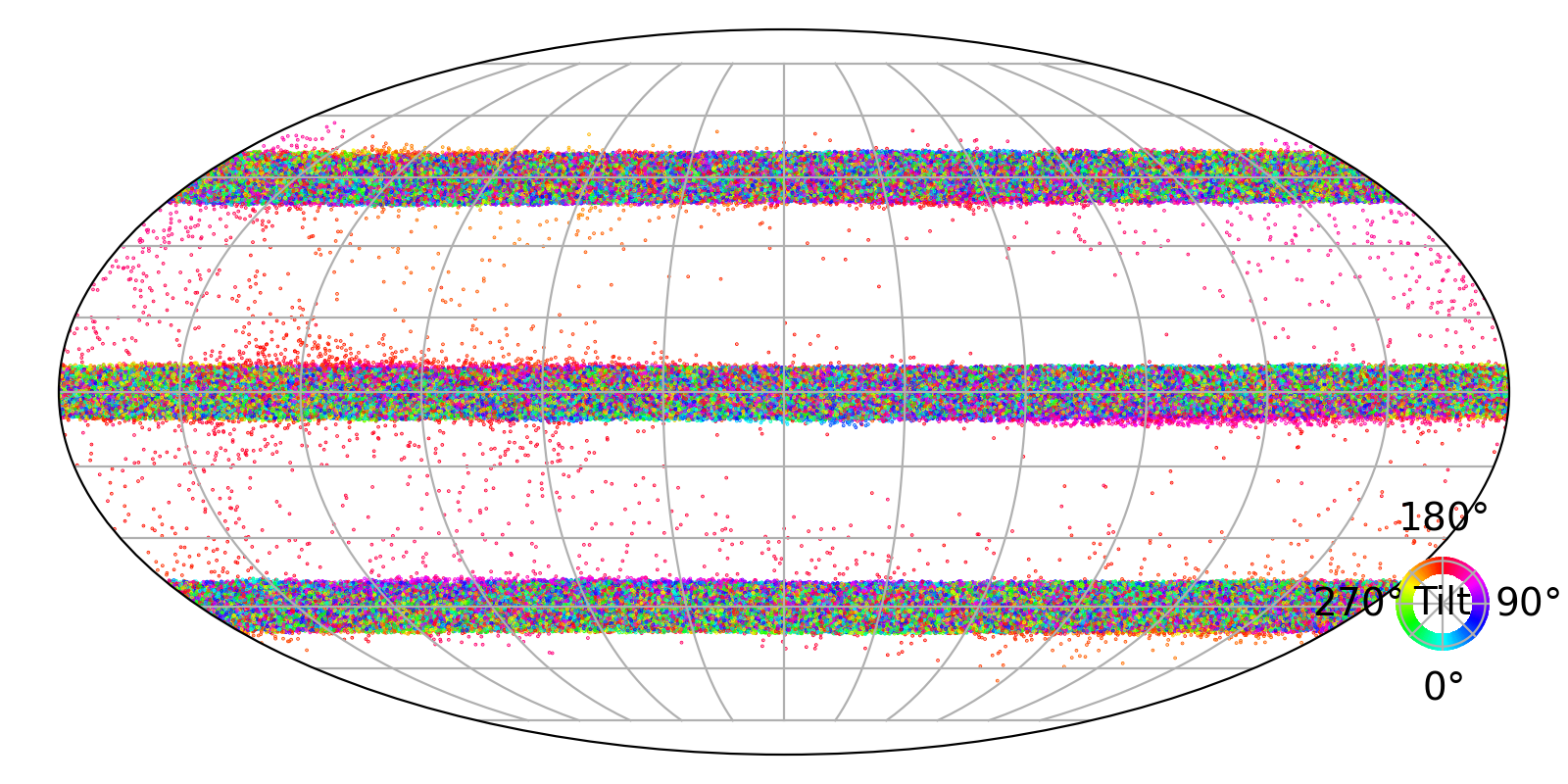}\\
    \includegraphics[width=0.23\linewidth]{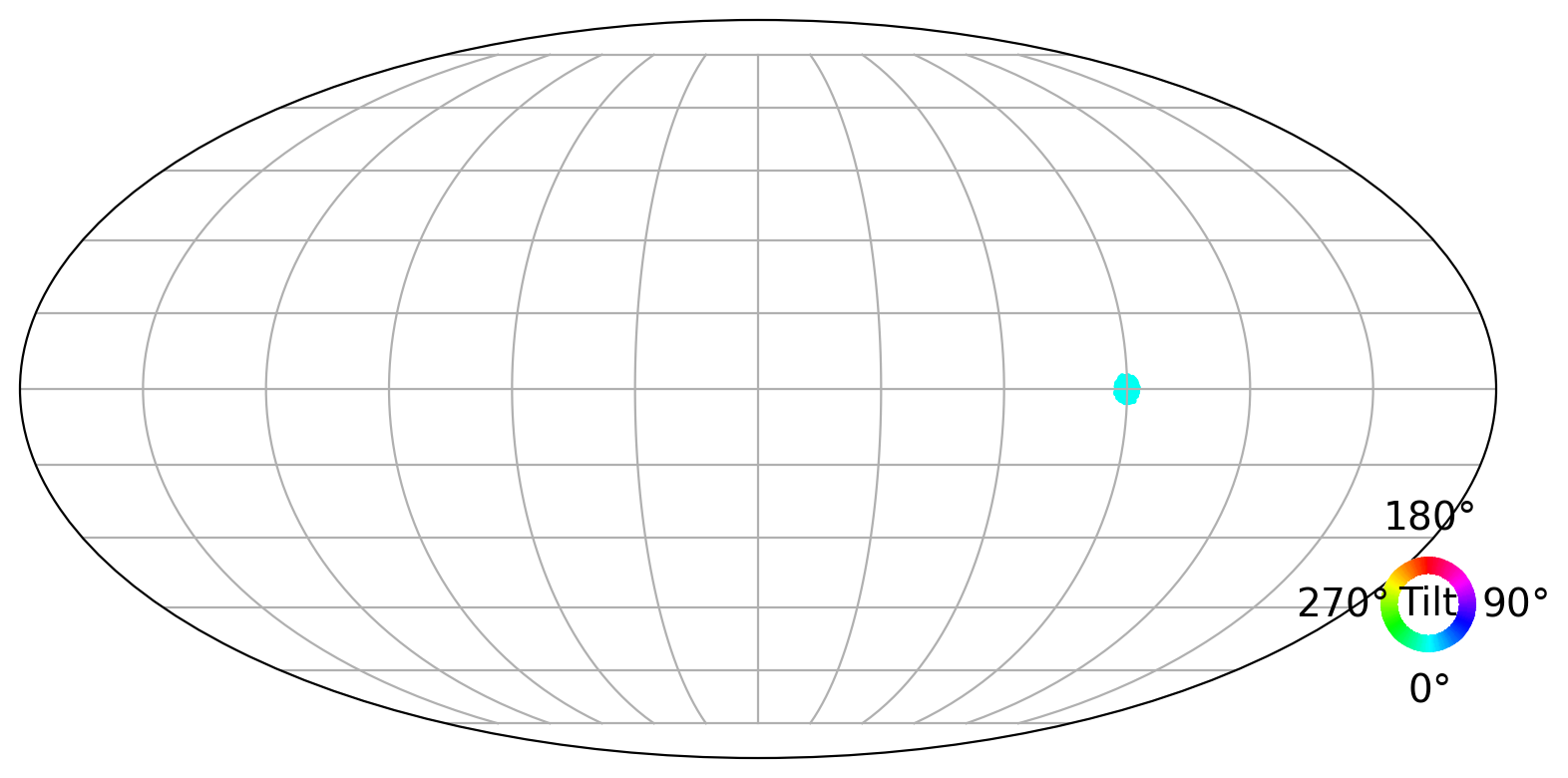} \hspace{0.5mm}
    \includegraphics[width=0.23\linewidth]{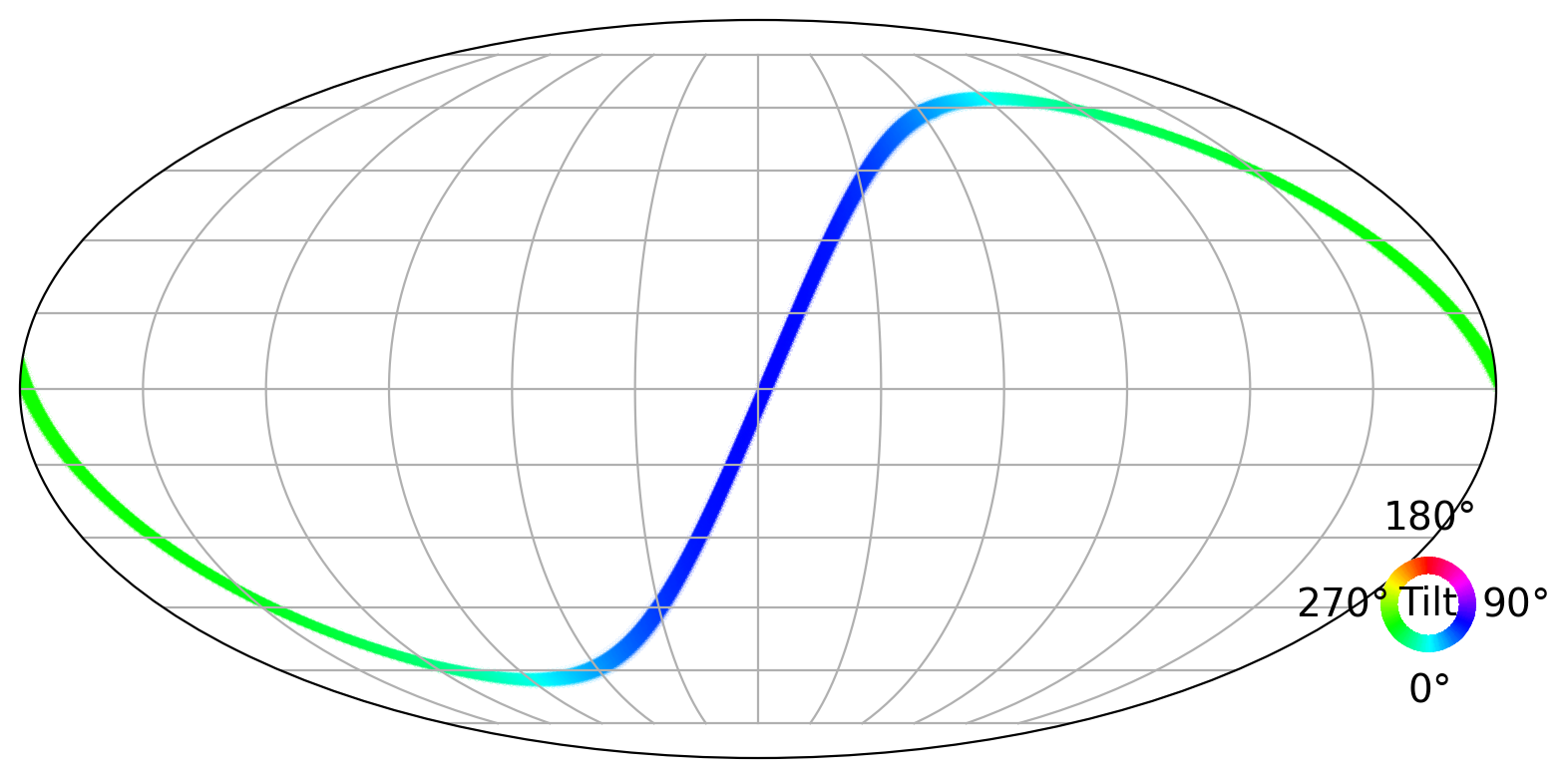} \hspace{0.5mm}
    \includegraphics[width=0.23\linewidth]{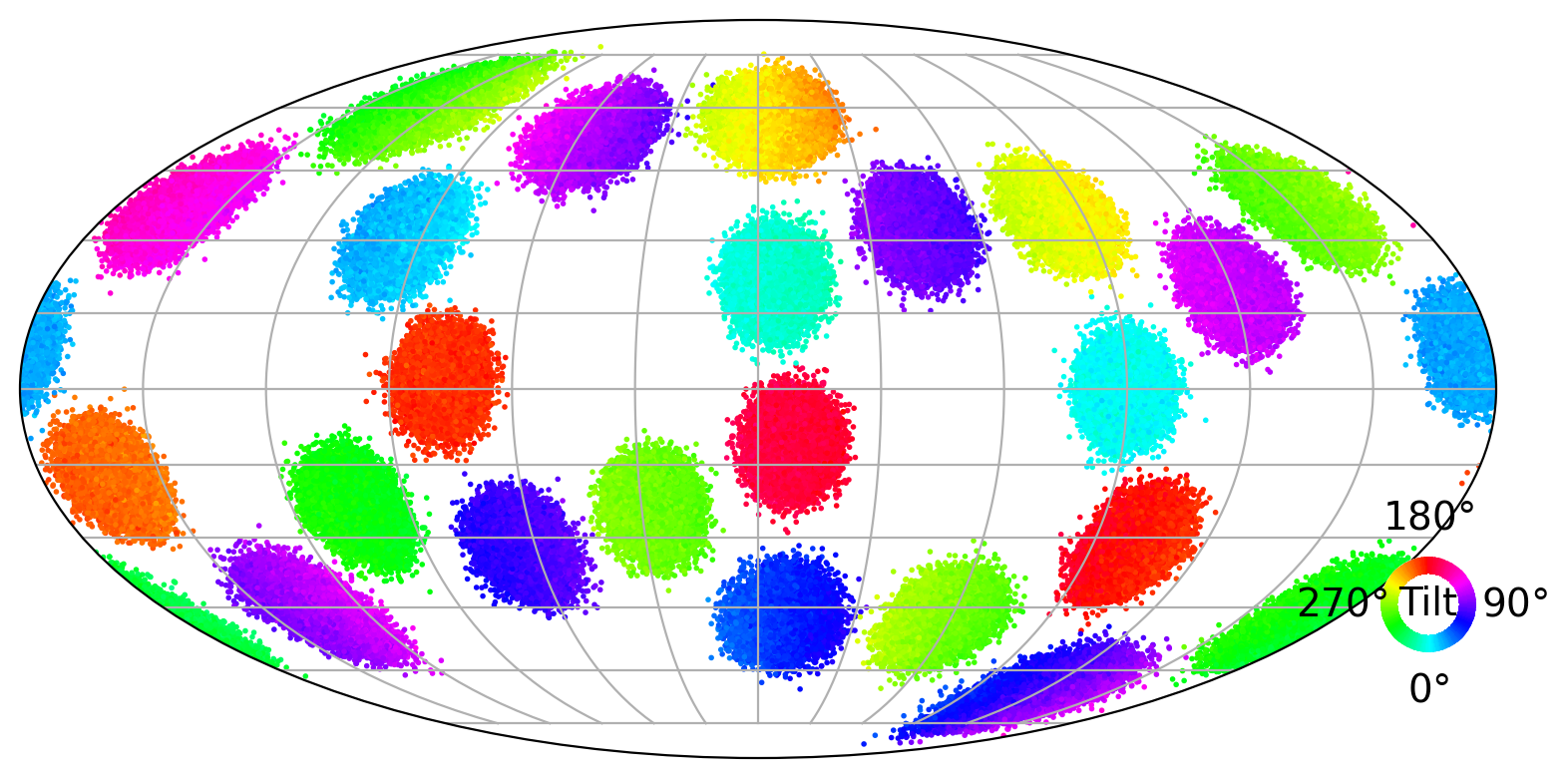} \hspace{0.5mm}
    \includegraphics[width=0.23\linewidth]{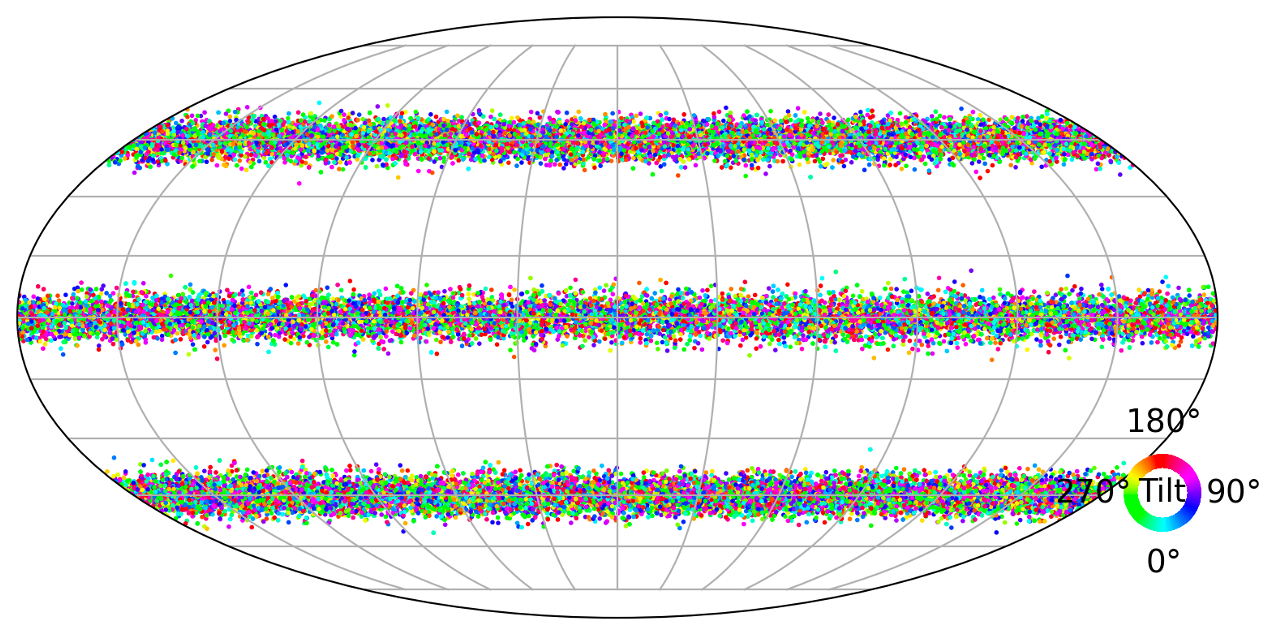}
    \caption{\textbf{Visualization of learned distributions for synthetic dataset}. \textbf{Top}: our learned distributions. \textbf{Bottom}: ground truth of density distributions. From the left to right are respectively peak, cone-like, cube-like, and line distributions. The distribution visualization is the same as Figure \ref{fig:pipeline}.}
    \label{fig:visualization}
\end{figure}

\subsection{Rotation Regression with Conditional Normalizing 
Flows}\label{sec: conditional}
In this experiment, we leverage our normalizing flows on conditional rotation regression given a single image. 
\subsubsection{SYMSOL I/II}
\noindent\textbf{Datasets}  We experiment on SYMSOL dataset introduced by \cite{murphy2021implicit}. SYMSOL I dataset contains images with solids with high order of symmetry, e.g., tetrahedron, cube, cone, cylinder, which challenges probabilistic
approaches to learn complex pose distributions. SYMSOL II dataset includes solids with small markers to break the regular symmetries.
We follow the experiment settings of \cite{murphy2021implicit}.

\noindent\textbf{Baselines}  We compare our method to Implicit-PDF \cite{murphy2021implicit} as well as several works which parameterize multimodal distributions on $\SO$ for the purpose of pose estimation, including von-Mises distribution \cite{prokudin2018deep} and Bingham distribution \cite{deng2022deep,gilitschenski2019deep}. We quote numbers of baselines from \cite{murphy2021implicit}.

\noindent\textbf{Results} 
The log-likelihood scores are reported in Table \ref{tab:symsol_loglik}. We can see that on both SYMSOL I and II datasets, our proposed rotation normalizing flows obtain a significant and consistent performance improvement over all the baselines.
We further evaluate our method under \textit{spread} metric in Table \ref{tab:spread}. \textit{Spread}, also referred to as the Mean Absolute Angular Deviation (MAAD)\cite{murphy2021implicit,gilitschenski2019deep,prokudin2018deep}, measures the minimum expected angular deviation to the equivalent ground truth $\mathbb{E}_{R\sim p(R|x)}[\min_{R'\in \{R_{GT}\}}d(R, R')]$ and $d(R,R')$ is the geodesic distance between rotations. Results show that our model learns to predict accurate and concentrated rotation pose, with an average MAAD less than 1$^\circ$. 
\begin{figure}
    \centering
    a \includegraphics[width=0.15\columnwidth]{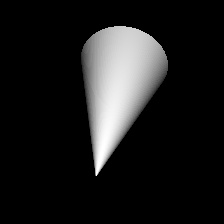}
    \includegraphics[width=0.3\columnwidth]{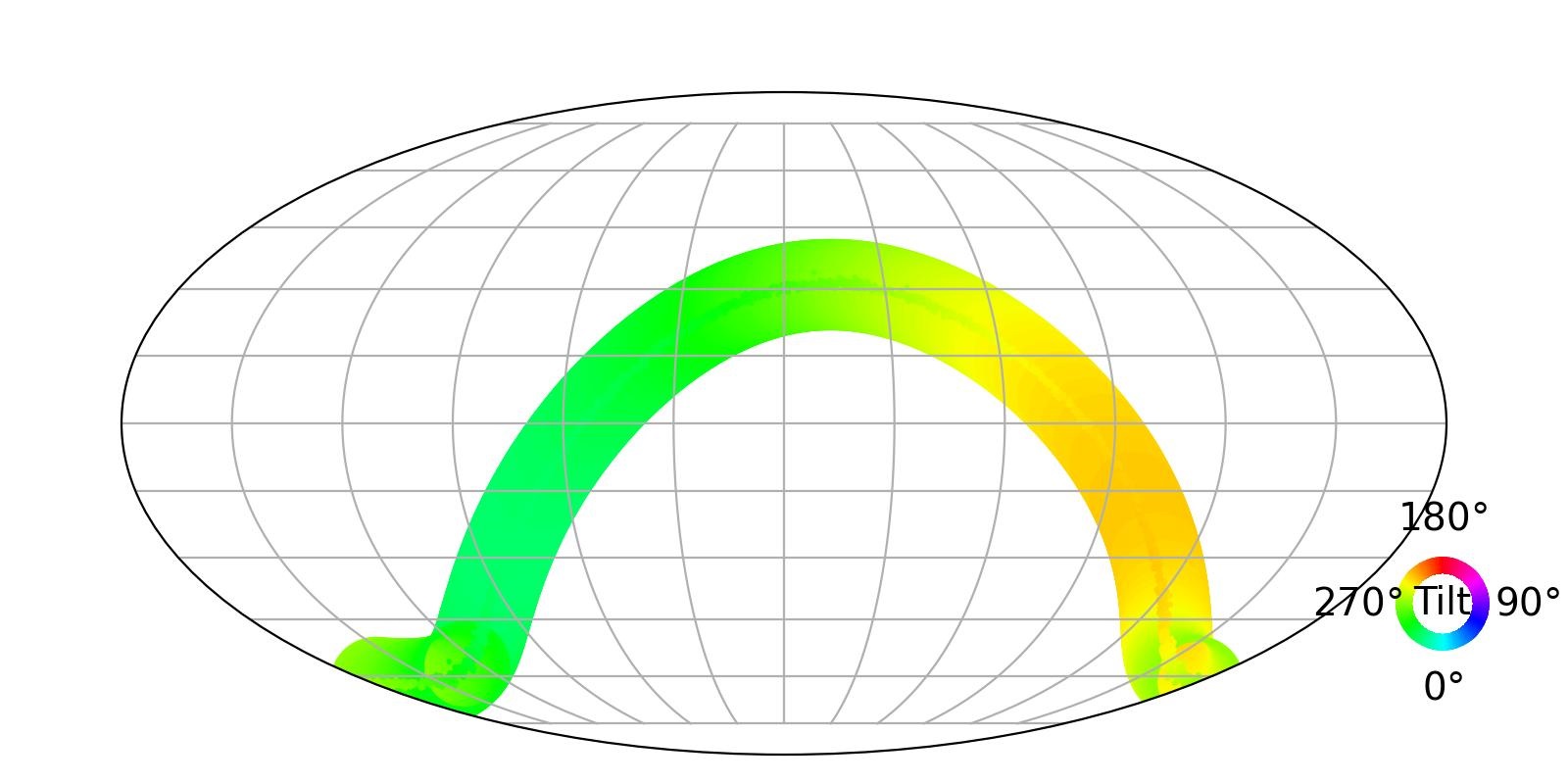}
    b \includegraphics[width=0.15\columnwidth]{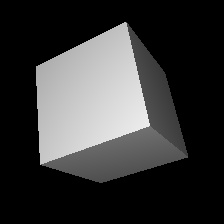}
    \includegraphics[width=0.3\columnwidth]{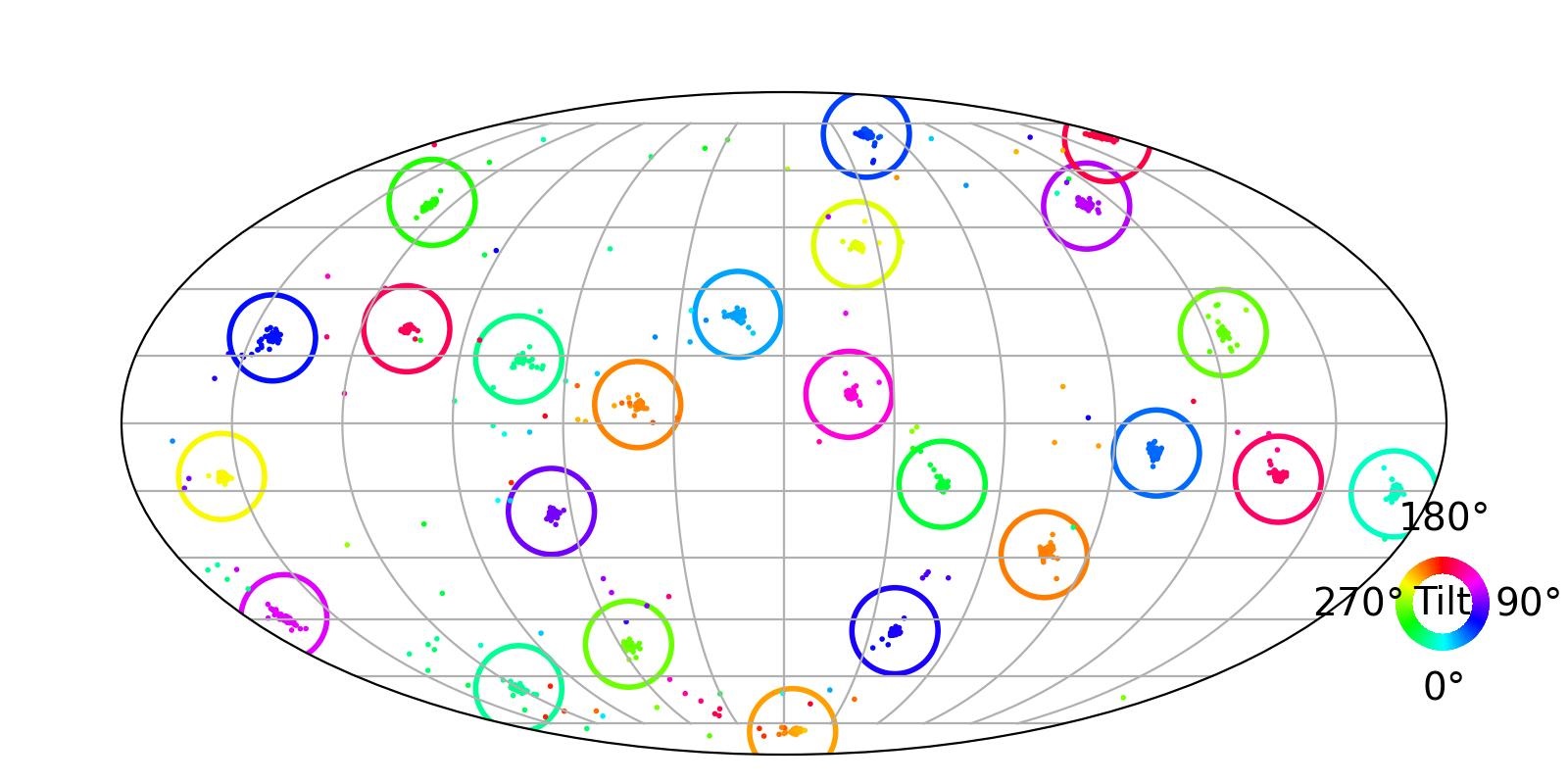}
    c \includegraphics[width=0.15\columnwidth]{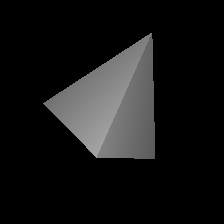}
    \includegraphics[width=0.3\columnwidth]{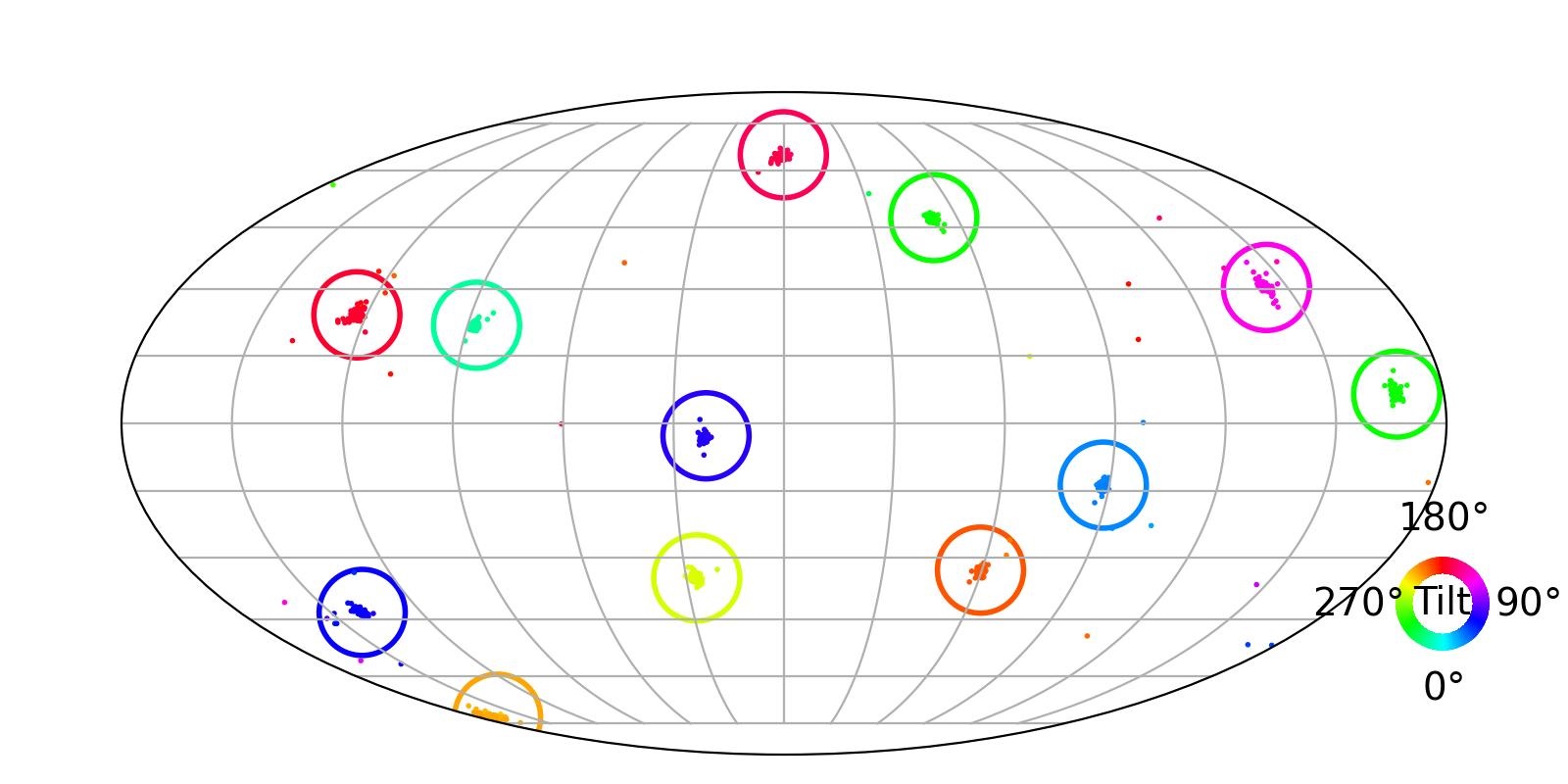}
    d \includegraphics[width=0.15\columnwidth]{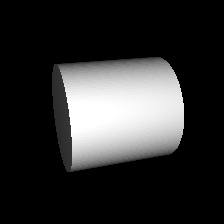}
    \includegraphics[width=0.3\columnwidth]{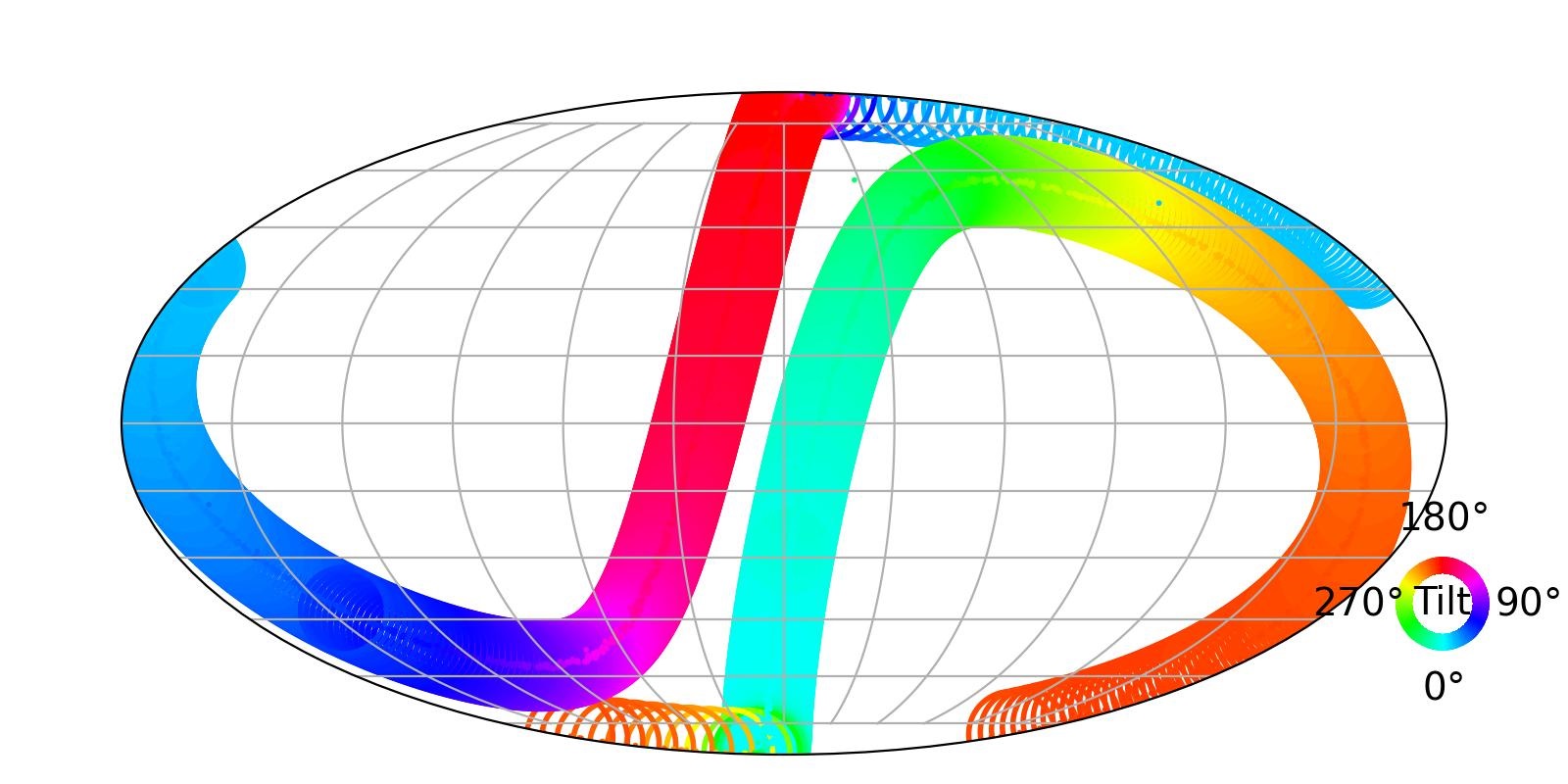}
    e \includegraphics[width=0.15\columnwidth]{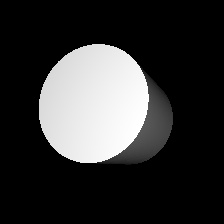}
    \includegraphics[width=0.3\columnwidth]{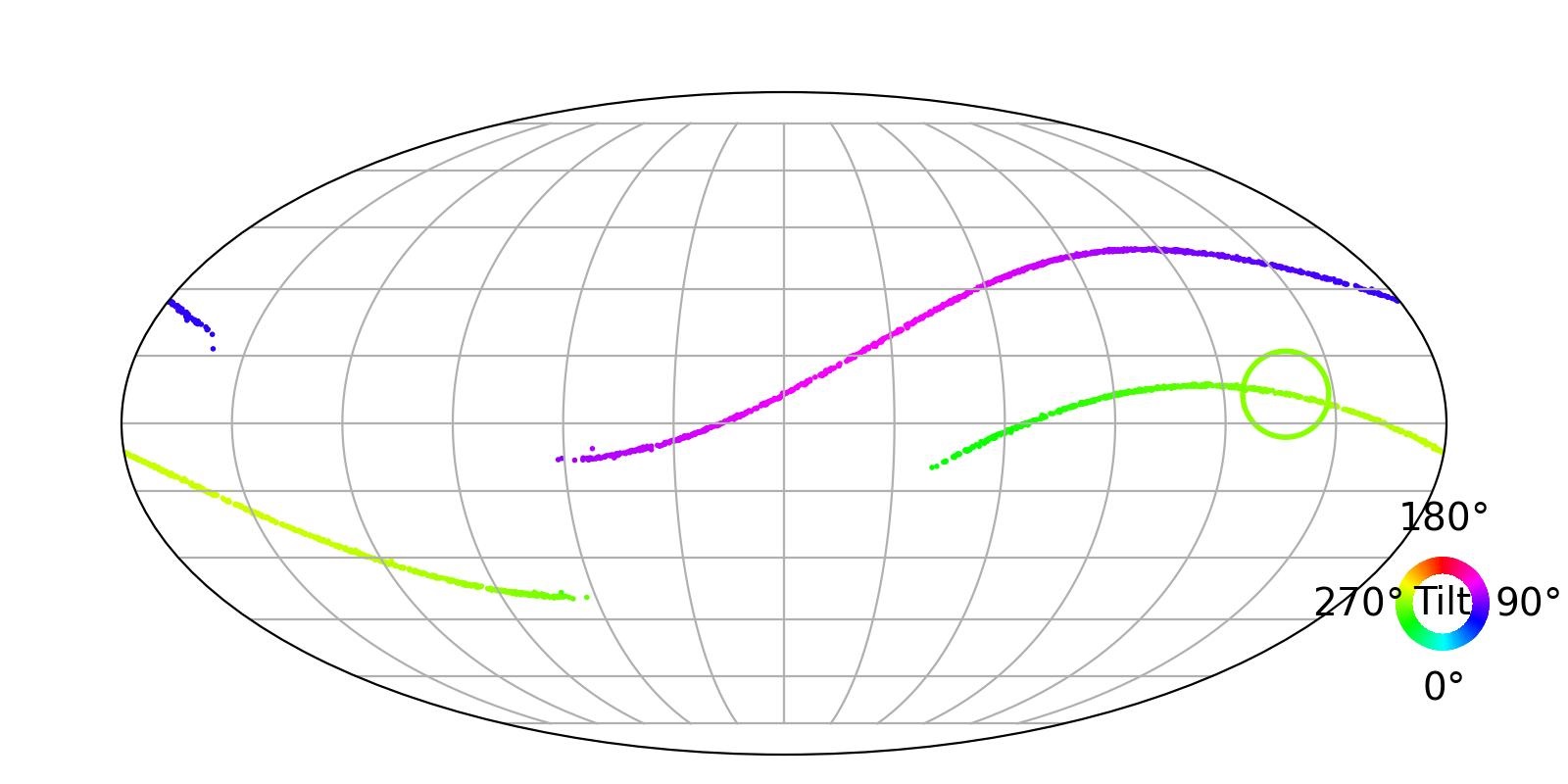}
    f \includegraphics[width=0.15\columnwidth]{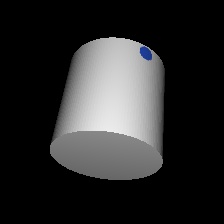}
    \includegraphics[width=0.3\columnwidth]{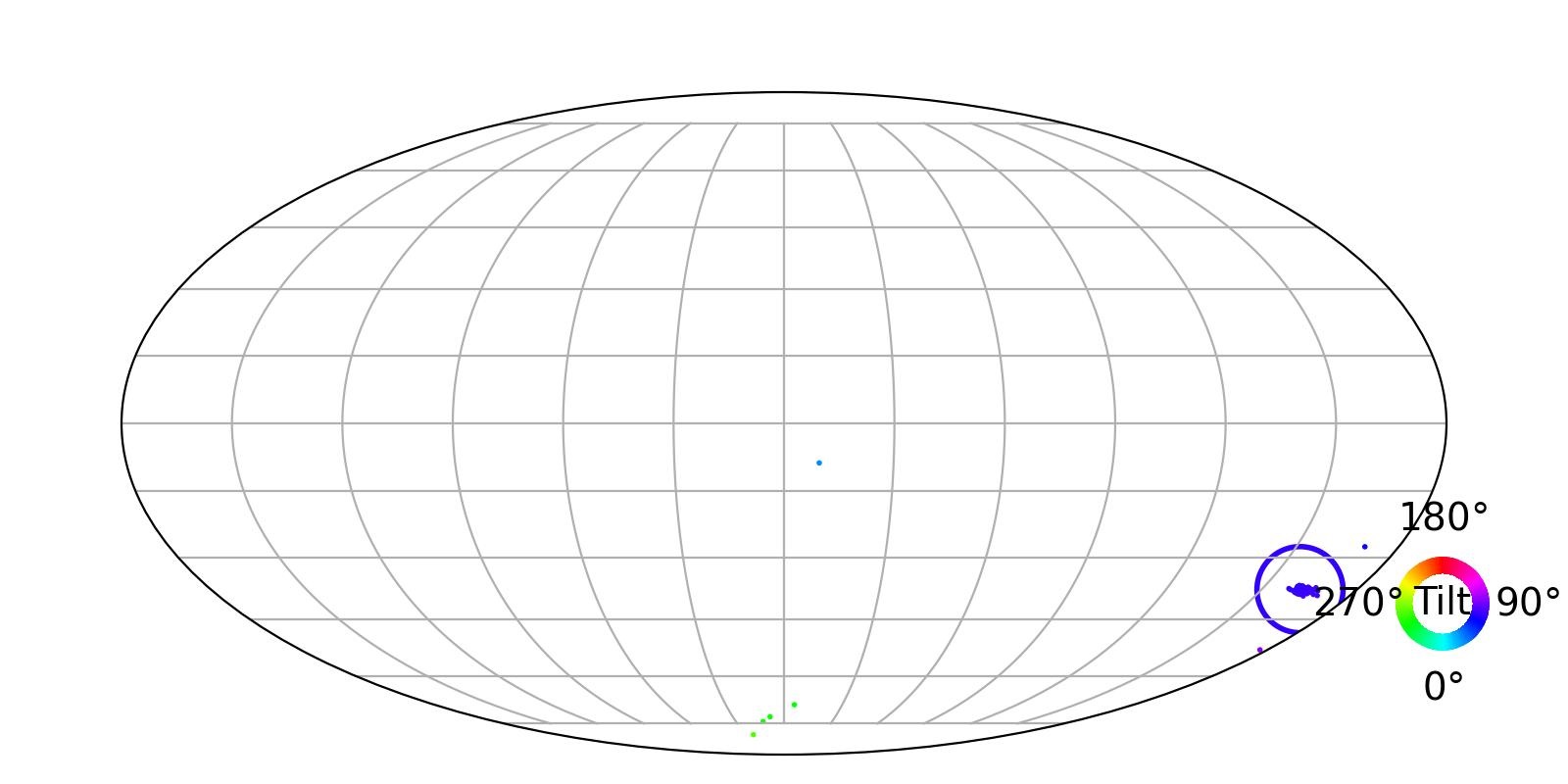}
    \caption{\textbf{Visualization of learned distributions for SYMSOL}. The ground truth is shown as a big circle around the point on the sphere which corresponds to the first axis and is colored according to the rotation of the axis. \textbf{a}: The cone possesses rotational symmetry and has a circle of equivalent poses. \textbf{b,c}: 24-modal distribution of cube and 12-modal distribution of tetrahedron are well fitted by our flow model. \textbf{d,e,f}:  (d)Cylinder without a mark has 2 circles of equivalent rotations under 2-fold rotation symmetry. (e,f)The cylinder is marked with a dot. When the dot is invisible, orientations with the mark on the hidden side are possible, thus our model predicts 2 half circles of rotations. When dot is visible, our model gives a single and accurate prediction.}
    \label{fig:visulaization2}
    \vspace{-3pt}
\end{figure}

\setlength{\tabcolsep}{9pt}
\begin{table*}[!hbt]
\centering
\scriptsize
\caption{
\textbf{Results of rotation regression with conditional normalizing flows on SYMSOL I and II.} We adopt log-likelihood as the evaluation metric and use uniform distribution in $\SO$ as base distribution. Note that we use the convention that a minimally informative uniform distribution has an average log likelihood of 0, which is different from IPDF's convention of -2.29.
}
\begin{tabular}{lccccc c ccccc}
  \toprule
    & \multicolumn{6}{c}{SYMSOL I (log likelihood $\uparrow$)}                           &           & \multicolumn{4}{c}{SYMSOL II (log likelihood $\uparrow$)}                                                                       \\
    \cmidrule{2-7}
    \cmidrule{9-12}
    & {avg.}                                                 & {cone}    & {cube}     & {cyl.}    & {ico.}     & {tet.}     &  & {avg.}    & {sphX}    & {cylO}    & {tetX}    \\
    \midrule
    Deng et al.\cite{deng2022deep}          & 0.81                                                 & 2.45  & -2.15    & 1.34    & -0.16   &  2.56            &   & 4.86     & 3.41      & 5.28      &  5.90 \\
    Gilitschenski et al. \cite{gilitschenski2019deep} &  1.86                                             & 6.13 & 0.00 &  3.17 & 0.00     &    0.00 &  &  5.99  &  5.61 &  7.17 & 5.19      \\
    Prokudin et al. \cite{prokudin2018deep}      & 0.42                                                 & -1.05  &  1.79    & 1.01     & -0.10   & 0.43         &  & 2.77     & -1.90     & 6.45     & 3.77      \\
    IPDF \cite{murphy2021implicit}     &  6.39                                               &  6.74&  7.10 &  6.55 & 3.57 &  7.99  &  &  9.86 &  9.59 &  9.20 &  10.78 \\
    \hline
    Ours     &  \textbf{10.38} & \textbf{10.05}  & \textbf{11.64}  & \textbf{9.54}   &\textbf{8.26} & \textbf{12.43}  &  & \textbf{12.94} & \textbf{12.37}  &  \textbf{12.92} & \textbf{13.53}  \\
  \bottomrule
\end{tabular}

\label{tab:symsol_loglik}
\end{table*}
\begin{table}[t]
  \centering
  \scriptsize
  \caption{\textbf{\textit{Spread} estimation on SYMSOL.} This metric evaluates how close the probability mass is centered on any of the equivalent ground truths. We follow \cite{murphy2021implicit} to evaluate it on SYMSOL I, where all ground truths are known at test time. Values are in degrees.}
    \begin{tabular}{lcccccc}
    \toprule
         \textit{spread$\downarrow$} & \multicolumn{1}{l}{avg.}& \multicolumn{1}{l}{cone} & \multicolumn{1}{l}{cube} & \multicolumn{1}{l}{cyl.} & \multicolumn{1}{l}{ico.}& \multicolumn{1}{l}{tet.} \\
    \midrule
    Deng et al.\cite{deng2022deep} &  22.4&  10.1    &  40.7     &   15.2    & 29.5 & 16.7 \\
    IPDF\cite{murphy2021implicit}  & 4.0 &   1.4   & 4.0 &   1.4    &    8.4 & 4.6 \\
    Ours  & \textbf{0.7} &   \textbf{0.5}    &   \textbf{0.6}    &   \textbf{0.5}    & \textbf{1.1}&\textbf{0.6}  \\
    \bottomrule
    \end{tabular}
  \label{tab:spread}
  \end{table}

\yulin{\subsubsection{ModelNet10-SO3}

\noindent\textbf{Dataset} ModelNet10-SO3 dataset \cite{liao2019spherical} is a popular dataset widely used in the task of regressing rotations from single images. It is synthesized by rendering the CAD models of ModelNet10 dataset \cite{wu20153d} that are uniformly rotated.

\noindent\textbf{Baselines} We have established our baselines using recent works on probabilistic rotation regression, including von Mises distribution~\cite{prokudin2018deep}, Bingham distribution~\cite{deng2022deep}, matrix Fisher distribution~\cite{mohlin2020probabilistic} and Implicit-PDF~\cite{murphy2021implicit}.

\noindent\textbf{Results} We leverage the common acc@15$^\circ$, acc@30$^\circ$ and median error metrics to evaluate the performance, and the detailed results are shown in Table \ref{tab:m10so3-pascal}. \haoran{We show results of using uniform distribution as base distribution (Ours(Uni.)) and a pre-trained fisher distribution as base distribution (Ours(Fisher)).}
As shown in the table, our method yields superior in the task of unimodal rotation regression. Note that the main advantage of our normalizing flows over other parametric distributions is the superiority to model complex distributions on $\SO$, however, the results of unimodal rotation regression further demonstrate the robustness and wide applications of our method.

\begin{table}[t]
\caption{
    \textbf{Numerical results of rotation regression on ModelNet10-SO(3).}
    We adopt $15^{\circ}$ and $30^{\circ}$ accuaracy and median error as evaluation metric. Metrics are averaged over categories, and the best performance is shown in \textbf{bold} while the second best is \underline{underlined}.
    See Supplementary Material for the complete table with per-category metrics.
  }
  \centering
    \scriptsize
    \begin{tabular}{@{}lccc}
      \toprule
 & {Acc@}{15\textdegree  $\uparrow$} & {Acc@}{30\textdegree  $\uparrow$} & \shortstack{Med.} \shortstack{\ ($^\circ$) $\downarrow$}\\
      \midrule
      Deng et al.\cite{deng2022deep}            
      & 0.562                           & 0.694                           & 32.6                                       \\
      Prokudin et al.\cite{prokudin2018deep}        
      & 0.456                           & 0.528                           & 49.3                                       \\
      Mohlin et al.\cite{mohlin2020probabilistic} 
      &  0.693                      & 0.757                      &  17.1                                  \\    
      IPDF \cite{murphy2021implicit}                    
      &  0.719                      &  0.735                      & 21.5                                  \\
      Ours (Uni.)
      & \textbf{0.760} & \textbf{0.774} &  \underline{14.6}\\
      Ours (Fisher)
      & \underline{0.744} & \underline{0.768} &  \textbf{12.2}\\
      
      \bottomrule
    \end{tabular}
    
  \label{tab:m10so3-pascal}
\end{table}

\begin{table}[ht]
  \centering
  \scriptsize
  \caption{\textbf{Numerical results of rotation regression on Pascal3D+ dataset.} We adopt $30^{\circ}$ accuaracy and median error as the evaluation metrics. Metrics are averaged over categories, and the best performance is shown in \textbf{bold}. See Supplementary Material for the complete table with per-category metrics.}
    \begin{tabular}{@{}lcc}
      \toprule
           & Acc@30$^{\circ}\uparrow$    & Med.($^\circ$)$\downarrow$\\  
      \midrule

   Liao et al. \cite{liao2019spherical}        &   0.819&       13.0  \\
    Mohlin et al. \cite{mohlin2020probabilistic}      &   0.825&       11.5  \\
    Prokudin et al.  \cite{prokudin2018deep}   &   0.838&       12.2  \\
    Tulsiani \& Malik \cite{tulsiani2015viewpoints}  &   0.808&       13.6  \\
     Mahendran et al. \cite{mahendran2018mixed}   &   \underline{0.859}&       \underline{10.1}  \\
    IPDF       \cite{murphy2021implicit}         &   0.837&       10.3  \\
    Ours (Uni.)     &  0.827 &     10.2    \\ 
    Ours (Fisher) & \textbf{0.863} & \textbf{9.9}\\
    \bottomrule
    \vspace{-4mm}
 
    \end{tabular}
  \label{tab:pascal}
\end{table}
}

\yulin{
\subsection{Pascal3D+}
\noindent\textbf{Dataset} Pascal3D+ dataset \cite{xiang2014beyond} is \haoran{also a popular dataset in rotation regression with a single image as input and it consists of real images with no symmetry.}

\noindent\textbf{Baselines} Our baselines consist of recent works for probabilistic rotation regression, including von Mises distribution~\cite{prokudin2018deep}, matrix Fisher distribution~\cite{mohlin2020probabilistic} and Implicit-PDF~\cite{murphy2021implicit} \haoran{ and non-probabilistic rotation regression methods including \cite{liao2019spherical}, \cite{tulsiani2015viewpoints} and \cite{mahendran2018mixed}}.

\noindent\textbf{Results} 
As shown in Table \ref{tab:pascal},
\haoran{by pretraining a neural network which uses Matrix Fisher distribution~\cite{mohlin2020probabilistic} to estimate rotation and then using the learned distribution as the base distribution in our flow, our method can achieve the highest acc@30$^\circ$ rate and the lowest median error. This is because the pre-trained conditional Matrix Fisher distribution can serve as a better initialization to our flow, and our flow can further improve the performance as it is able to model more complex distributions and can thus approximate the underlying distribution better. This result shows that our method can collaborate with other probabilistic methods to get better performance. }
}
\subsection{Ablation Study}
\label{sec:ablation}
In this experiment, we evaluate the effectiveness of each proposed component in our rotation normalizing flows. To show the 2 functions of affine transformation, we implement Mobius with Rotation only, where $W$ of affine transformation is replaced by $R=UV^T$ of its SVD decomposition $W=USV^T$. The experiment settings are the same as Sec. \ref{sec: conditional} and the results are reported in Table \ref{tab:ablation}. 
\par
As shown in Table \ref{tab:ablation}, Mobius coupling layers are of crucial importance while only quaternion affine transformation is not capable to tackle such complex distributions. Quaternion affine transformation performs as an enhancement to Mobius coupling layers and accelerates learning. 
\par
\begin{figure}
    \centering
    \includegraphics[width = 0.48\columnwidth]{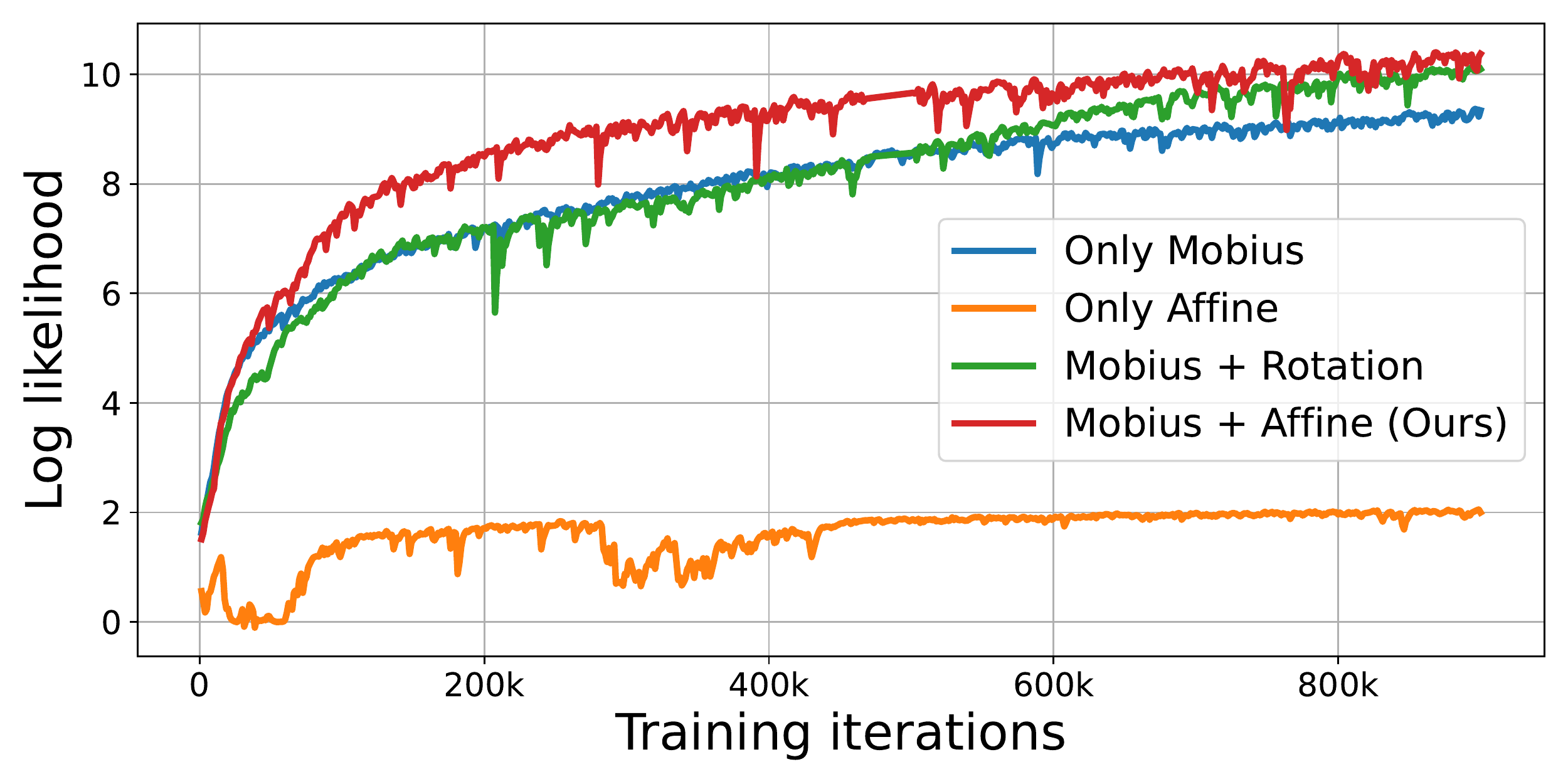}
    \includegraphics[width = 0.48\columnwidth]{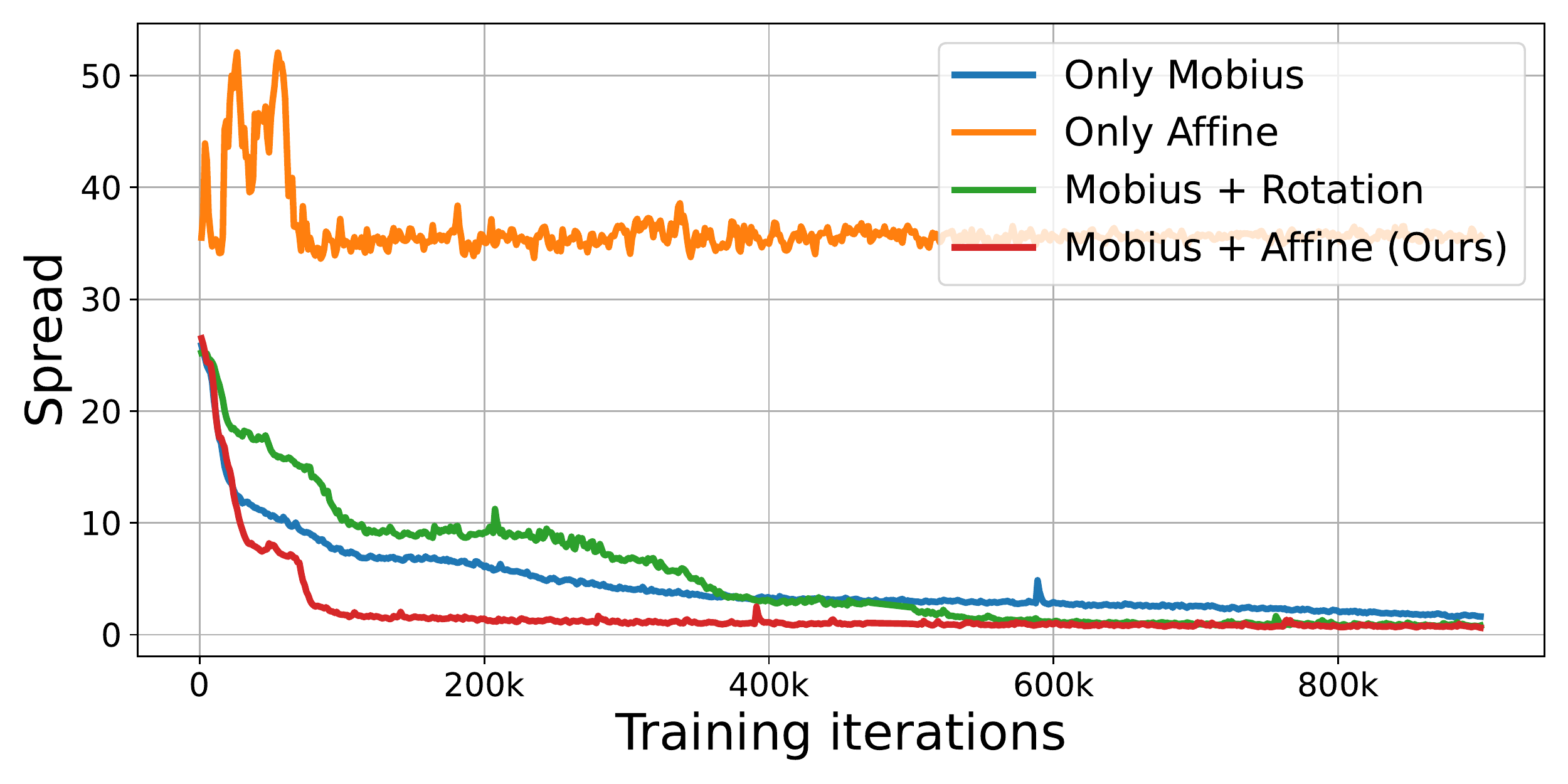}
    \caption{\textbf{Ablation study on Affine transformation on SYMSOL I dataset}. We plot \textit{log likelihood} and \textit{spread} evolving with training iterations. }
    \vspace{-6mm}
    \label{fig:ablation}
\end{figure}

For \textit{spread}, as illustrated in Figure \ref{fig:ablation}, it takes Mobius with affine about 100k iterations for convergence, whereas Mobius without affine continued to decrease for 900k iterations. Comparison between Mobius with affine and Mobius with rotation shows the effects of scaling and normalization \haoran{part} to accelerate learning speed as it can quickly concentrate distributions and approximate high mode.
\par
Though \textit{spread} quickly converges, the log-likelihood continues to increase, as the confidence of predictions is enhanced \haoran{during} 
training.

\begin{table}[tb]
\captionof{table}{\textbf{Ablation study on Affine transformation on SYMSOL I dataset}. We adopt \textit{log-likelihood} and \textit{spread} as evaluation matrice. }

    \centering
    \scriptsize
    \resizebox{\linewidth}{!}{
    \begin{tabular}{lcccccc}
    \toprule
          (log likelihood $\uparrow$) & \multicolumn{1}{l}{avg.} & \multicolumn{1}{l}{cone} & \multicolumn{1}{l}{cube} & \multicolumn{1}{l}{cyl.} & \multicolumn{1}{l}{ico.} &
          \multicolumn{1}{l}{tet.} \\
    \midrule
    Mobius & 9.41& 10.52  &  9.68     &   \textbf{10.00}    &   5.35    &11.51   \\
    Affine & 1.96& 9.79 &   0.00    &    0.00  &    0.00 & 0.00  \\
    Mobius + Rotation & 10.06& \textbf{10.65} & 9.91   & 9.99     &   7.97  & 11.78 \\
    Mobius + Affine (Ours) & \textbf{10.38} &   10.05    &  \textbf{11.64}    &  9.54    &  \textbf{8.26}&\textbf{12.43}\\
    \bottomrule
    \end{tabular}
    }
\vspace{0.2cm}

    \centering
    \scriptsize
    \resizebox{\linewidth}{!}{
    \begin{tabular}{lcccccc}
    \toprule
          (\textit{spread} $\downarrow$)& \multicolumn{1}{l}{avg.} & \multicolumn{1}{l}{cone} & \multicolumn{1}{l}{cube} & \multicolumn{1}{l}{cyl.} & \multicolumn{1}{l}{ico.} &
          \multicolumn{1}{l}{tet.} \\
    \midrule
    Mobius & 1.60&0.45&1.00&\textbf{0.42}&5.18&0.96\\
    Affine & 35.49&0.51&41.16&56.45&29.05&50.29\\
    Mobius + Rotation&0.87&\textbf{0.40}&1.63&0.43&\textbf{1.06}&0.84\\
    Mobius + Affine(Ours)&\textbf{0.67}&0.50&\textbf{0.60}&0.53&1.08&\textbf{0.64}\\
    \bottomrule
    \end{tabular}
    }\label{tab:ablation}
\end{table}

\section{Conclusion}
In this work, we show the capability of our proposed novel discrete normalizing flows for rotations to learn various kinds of distributions on $\SO$. The proposed flow is numerically stable and very expressive, thanks to the complementary role of our proposed Mobius coupling layer and quaternion affine transformation. Our extensive experiments demonstrate that our flows are able to fit complex distributions on $\SO$ and achieve the best performance in both unconditional and conditional tasks. 
\par
The Mobius coupling can be elegantly interpreted as a bundle isomorphism. It uses the fiber bundle with projection $\pi: \SO \rightarrow S_2$ and fiber $S_1 = SO(2)$. Choosing the conditional dimension $i = 1, 2, 3$  defines one such projection, for each base space point, there is a diffeomorphism over its fiber, which together form a base-space preserving bundle isomorphism. Our Mobius coupling design and affine transformation for accelerating may shed light on normalizing flows for other fiber bundles.  

\section*{Acknowledgement}
This work is supported in part by National Key R\&D Program of China 2022ZD0160801.

{\small
\bibliographystyle{ieee_fullname}
\bibliography{egbib}
}


\appendix



\section{Why perform affine transformation on quaternion?}

It seems straightforward to perform affine transformation on the rotation matrix, however, we find that these methods concentrate the distribution in an undesired way. The simplest way is to left or right multiply a $3\times 3$ invertible matrix $W$ to rotation matrix $R$, and then use SVD decomposition or the Gram-Schmidt orthogonalization to project the result $WR$ or $RW$ into a rotation matrix, i.e. SVD($WR$), SVD($RW$), GS($WR$), GS($RW$). 

\begin{figure}
    \centering
    \vspace{1pt}
    \includegraphics[width=\columnwidth]{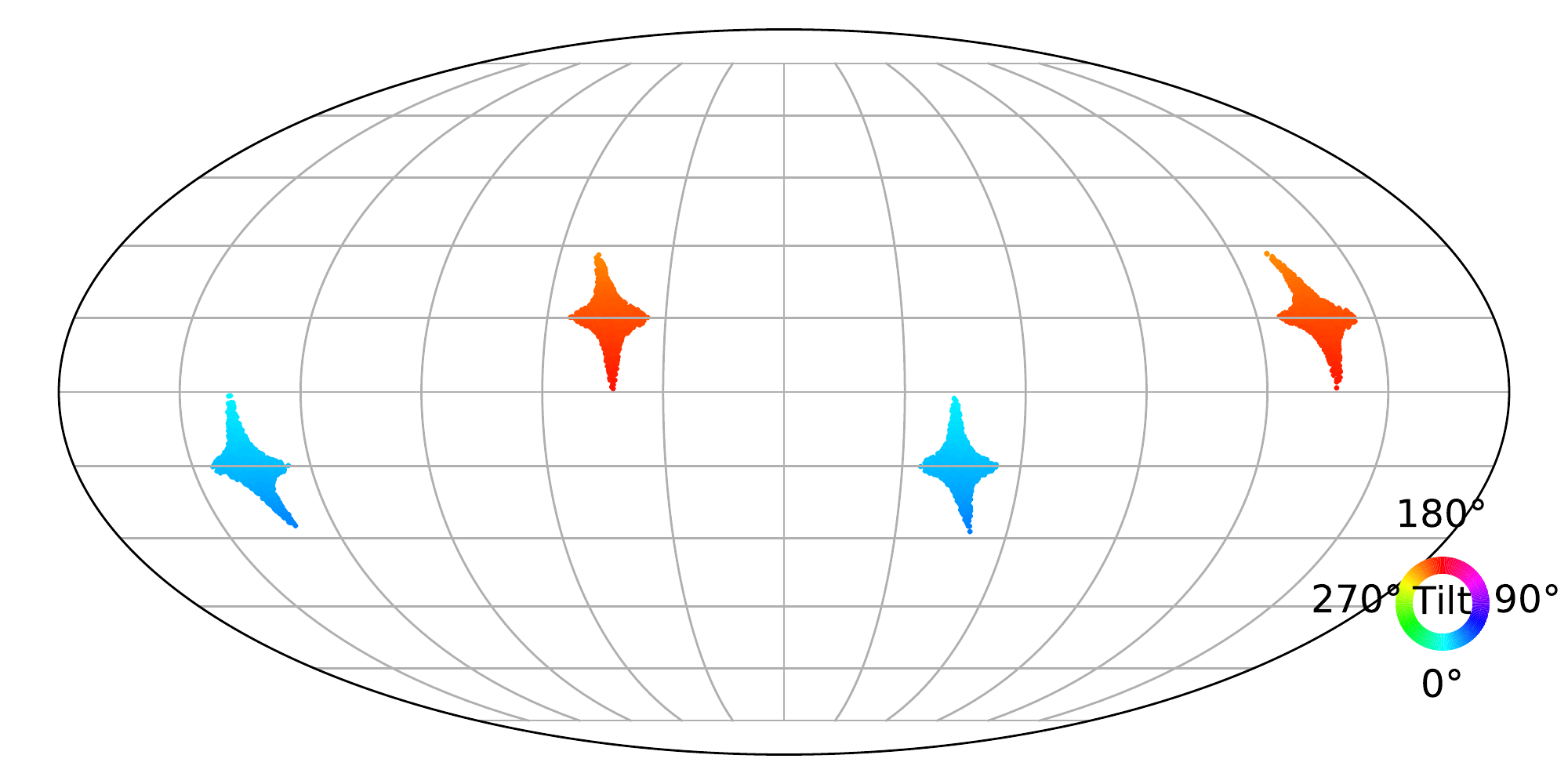}
    \caption{\textbf{Illustration of rotation matrix affine transformation.} It can be shown that when the rotation matrix affine transformation is performed by left multiplying $3\times 3$ invertible matrix W to the rotation matrix and then using the Gram-Schmidt orthogonalization. The probability is concentrated to 4 different modes, which usually leads to unsatisfactory performance. }
    \vspace{-4mm}
    \label{fig:3-3}
    \vspace{1pt}
\end{figure}

However, if we use SVD decomposition (SVD($WR$) or SVD($RW$)) or right Gram-Schmith orthogonalization (GS($RW$), it can be proved that it only functions as a rotation and thus has poor expressivity. If we use left Gram-Schmidt orthogonalization (GS($WR$)), we concentrate the distribution to $4$ different modes where the relative angle between each pair of them is fixed to be $180^\circ$ as shown in Figure \ref{fig:3-3}, which is usually undesired.

The 4-modal distribution can be understood as multiplying $W$ to the first column and then normalizing it can be interpreted as scaling $\mathrm{S}^2$ to an ellipsoid and then squeezing it to make it concentrated similar to quaternion affine transformation. Multiplying $W$ to the second column and using the Gram-Schmidt orthogonalization can be viewed as scaling a circle $\mathrm{S}^1$ to an oval and then squeezing it to make it concentrated. Those two operations both have the property of antipodal symmetry similar to quaternion affine transformation. However, keeping antipodal symmetry is needed in quaternion affine transformation as $q$ and $-q$ represent the same rotation, while in rotation matrix affine transformation, if $(c_1, c_2, c_3)$ is one mode where $c_i$ is the i-th column of the rotation matrix, then $(-c_1, -c_2, c_3)$, $(-c_1, c_2, -c_3)$ and $(c_1, -c_2, -c_3)$ will also be the mode due to this symmetry.

One can also treat the first two or all three columns of the rotation matrix as a vector and then perform affine transformation on it, but these methods usually lead to even more modes and the inverse of those transformations are much harder to solve.

On the contrary, as shown in Figure \ref{fig:affine}, the affine transformation on quaternion has elegant geometric interpretations and it can concentrate the distribution to only one mode, 
so we choose to perform affine transformation on quaternion.

Table \ref{tab:modelnet-abla} reports results of abaltion study of quaternion affine transformation and rotation matrix affine transformation on ModelNet10-SO3 datasets. It shows that the overall performance of quaternion affine transformation is better than rotation matrix affine transformation. Note that our average median error is higher because the result is highly affected by \textit{bathtub} category, whereas \textit{bathtub} exhibits symmetry while there is only one ground truth annotation for each image.

\begin{table*}[ht]
  \centering
  \resizebox{0.9\textwidth}{!}{
    \begin{tabular}{@{}llccccccccccc}
      \toprule
       &                                      & {avg.}     & {bathtub}  & {bed}      & {chair}    & {desk}     & {dresser}  & {tv}       & {n. stand} & {sofa}     & {table}    & {toilet}   \\
      \midrule
      
      \multirow{4}{*}{Acc@15\textdegree}
      &RSVD&0.724	&\textbf{0.430}	&0.861	&0.888	&0.613	&0.716	&0.790	&0.577	&0.888	&0.522	&0.952\\
    &LSVD&\underline{0.734}	&0.353	&0.872	&\underline{0.899}	&\underline{0.660}	&\underline{0.730}	&\underline{0.804}	&\underline{0.601}	&\underline{0.927}	&\underline{0.526}	&\underline{0.965}\\
    &RSmith&0.717	&0.400	&0.841	&0.866	&0.622	&0.710	&0.792	&0.588	&0.882	&0.524	&0.951\\
    &LSmith &0.725	&0.400	&\underline{0.873}	&0.875	&0.613	&0.722	&0.799	&0.597	&0.913	&0.506	&0.950\\

    & Ours & \textbf{0.760} &	\underline{0.402} & \textbf{0.896} & \textbf{0.927}	& \textbf{0.704} & \textbf{0.753} & \textbf{0.843} & \textbf{0.602} & \textbf{0.939} & \textbf{0.561}	& \textbf{0.975}\\

      \midrule
      \multirow{4}{*}{Acc@30\textdegree}                   
   &RSVD & 0.731& \textbf{0.439}&	0.863	&0.907	&0.618&	0.719	&0.817	&0.580	&0.894	&0.525	&0.954\\
    &LSVD &\underline{0.750}	&0.371	&\underline{0.880}	&\underline{0.926}	&\underline{0.678}	&\underline{0.742}	&\underline{0.841}	&\underline{0.615}	&\underline{0.934}	&\underline{0.535}	&\underline{0.972}\\
    &RSmith&0.726	&0.407	&0.843	&0.887	&0.626	&0.715	&0.820	&0.593	&0.884	&0.526	&0.956\\
    &LSmith&0.738	&0.411	&\underline{0.880}	&0.900	&0.632	&0.729&0.826	&0.605	&0.920	&0.516	&0.959\\
    & Ours & \textbf{0.774} &	\underline{0.419}	& \textbf{0.904} & \textbf{0.946} & \textbf{0.722} & \textbf{0.766} & \textbf{0.868}	& \textbf{0.617} & \textbf{0.948}	& \textbf{0.567} & \textbf{0.982}\\

      \midrule
      \multirow{4}{*}{\shortstack{Median \\ Error ($^\circ$)}} 
    &RSVD & \textbf{11.3}	& \textbf{91.4}	& \textbf{1.4} & \textbf{2.8}	& \underline{2.6}	& \textbf{1.3} &\underline{2.8}&\underline{1.9} & \textbf{1.4} & \underline{5.6}	& \textbf{1.9}\\
    &LSVD & \underline{12.2}	& \underline{93.0} &1.7	&3.1	&3.2 &1.8 &3.0 & 2.8 & 1.7 &8.9	&2.3\\
    &RSmith & 16.9 &146.8&\underline{1.5} & \underline{2.9}	&\textbf{2.5}	&\underline{1.4}	&\underline{2.8}	&\textbf{1.8}	&\textbf{1.4}	&5.9	&\underline{2.0}\\
    &LSmith & 13.5 &106.5&\underline{1.5}&3.1	&2.9&1.5&\underline{2.8}&2.2&\underline{1.5}&11.3&2.2\\
    & Ours & 14.6	& 124.8	& \underline{1.5} & \textbf{2.8} & 2.7 &	1.5 &\textbf{2.6} &	2.4 & \underline{1.5}	& \textbf{3.9} & \underline{2.0}\\

      \bottomrule
    \end{tabular}}
  \caption{\textbf{Ablation of different affine transformation on ModelNet10-SO3 dataset.} We adopt 15$^{\circ}$ accuaracy, $30^{\circ}$ accuaracy and median error as the evaluation metrics. We use uniform distribution in $\SO$ as base distribution. The best performance is shown in \textbf{bold} and the second best is with \underline{underlined}.}
  \label{tab:modelnet-abla}
\end{table*}

\section{How to obtain the $4\times4$ Invertible Matrix?}
\label{sec:supp-lu}
We present two methods to parameterize the invertible matrix $W$ in quaternion affine transformation. One is to output an unconstrained matrix, and the other is to obtain the matrix by LU decomposition $W=PL(U+S)$, as in \cite{kingma2018glow}. For this method, we follow \cite{kingma2018glow} to construct $P$ as a fixed permutation matrix, $L$ as a lower triangular matrix with ones on the diagonal, $U$ as an upper triangular matrix with zeros on the diagonal, and $S$ as a diagonal matrix with positive entries.

We conduct an ablation study on different strategies in both unconditional and conditional experiments, and the results are shown in Table \ref{tab:lu}. We find that $4\times4$ unconstrained matrix outperforms LU decomposition as shown in the table.
This phenomenon may result from the less expressivity of the construction by LU decomposition. Given $W=PL(U+S)$, 
the sign of the diagonal elements of $S$ is always positive, so it can only represent a subspace of 4$\times$4 invertible matrix, which limits the expressivity. For example, if $P$ is fixed to be the identity matrix, the rotation matrix $diag(-1, -1, 1, 1)$ can't be parameterized via this strategy.

\setlength{\tabcolsep}{6.5pt}
\setlength{\tabcolsep}{6.5pt}
\begin{table}[!h]
\centering
\footnotesize
\caption{
\textbf{Abalations on parameterization strategies of the $4\times 4$ invertible matrix}. We report results on synthetic datasets and SYMSOL-I dataset with log-likelihood ($\uparrow$) as the evaluation metric. We also report results on ModelNet10-SO3 dataset and adpot acc@15, acc@30 and error median as the evaluation metrics.
}
\begin{tabular}
{lccccc}
  \toprule

   Synthetic datasets  & {avg.}  & {peak} &{cone}     & {cube}    & {line}  \\  
    \midrule

    Ours (unconstrained $M$)
    & \textbf{7.23} & \textbf{13.93} &  \textbf{8.99} &  \textbf{4.81}&  \textbf{1.38} \\

    Ours (LU)&\textbf{7.23} & 13.92 & \textbf{8.99} & \textbf{4.81}& \textbf{1.38}\\
\end{tabular}
  \resizebox{\columnwidth}{!}{
    \begin{tabular}{lcccccc}
    \toprule
         {SYMSOL I} & \multicolumn{1}{l}{avg.}& \multicolumn{1}{l}{cone} & \multicolumn{1}{l}{cube} & \multicolumn{1}{l}{cyl.} & \multicolumn{1}{l}{ico.}& \multicolumn{1}{l}{tet.} \\
    \midrule
    Ours (unconstrained M) & \textbf{10.38} & \textbf{10.05}   & \textbf{11.64}  & \textbf{9.54}   & \textbf{8.26} & \textbf{12.43}\\
    Ours (LU) & 
    8.57 & 9.95 & 8.60 & 9.38 & 3.94 & 10.99\\

    \bottomrule
    \end{tabular}
    }
\begin{tabular}
{lccc}
  \toprule

   ModelNet10-SO3   & {acc@15$\uparrow$} & {acc@30$\uparrow$}     & {Median Error$\downarrow$} \\  
    \midrule
 
    Ours (unconstrained $M$)& \textbf{0.760} & \textbf{0.774} & \textbf{14.6}  \\
    Ours (LU)& 0.727 & 0.738 & 16.7 \\
    \bottomrule
\end{tabular}
\label{tab:lu}
\end{table}

\section{Ablation $\frac{\sqrt{2}}{2}$ trick}

\begin{figure}
    \centering
    \includegraphics[width=\linewidth]{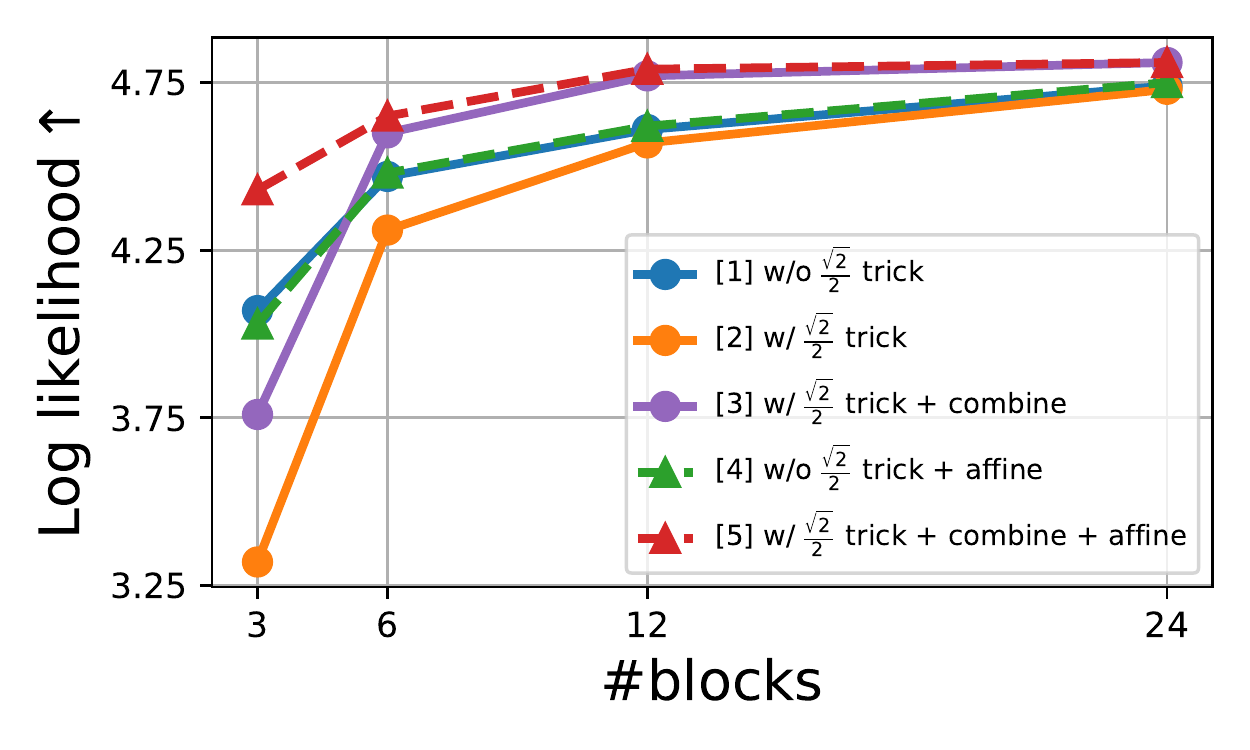}
    \caption{\textbf{Abalations on $\frac{\sqrt{2}}{2}$ trick on Cube dataset.} We plot results of log likelihood with different blocks and choice of structure. Curve[1] stands for single Mobius transformation without trick, [2] for single Mobius transformation with trick, [3] for 64-combination Mobius transformation with trick, [4] for single Mobius transformation without trick with affine transformation and [5] for 64-combination Mobius transformation with trick with affine transformation.}
    \label{fig:sqrt2-ablation}
\end{figure}
We present a $\frac{\sqrt{2}}{2}$ trick to alleviate discontinuity encountered in \textit{linear combination} of Mobius transformations. We have to confess that the expressivity of single Mobius transformation are reduced due to restriction of $\Vert \omega \Vert$, however, as it enables \textit{linear combination} of Mobius transformations, the general expressivity are gained. 

Without $\frac{\sqrt{2}}{2}$ trick, we can not compute the inverse process of \textit{linear combination} of Mobius transformations due to the ambiguity of Mobius combination. 
So we have to compare the gain from using linear combination and the cost of this $\frac{\sqrt{2}}{2}$ trick.

We conduct ablation study of $\frac{\sqrt{2}}{2}$ trick on Cube dataset in Fig. \ref{fig:sqrt2-ablation}. Seen from curves [1-2], the introduction of $\frac{\sqrt{2}}{2}$ trick indeed limits the performance; 
however, from [3] vs. [1] and [5] vs. [4], the accommodation of linear combination in return provides larger expressivity.

\section{More Discussions on using other Normalizing Flows on SO(3)}

\subsection{ReLie}
\label{sec:relie}
\begin{figure}
    \centering
    \includegraphics[width = 0.45\columnwidth]{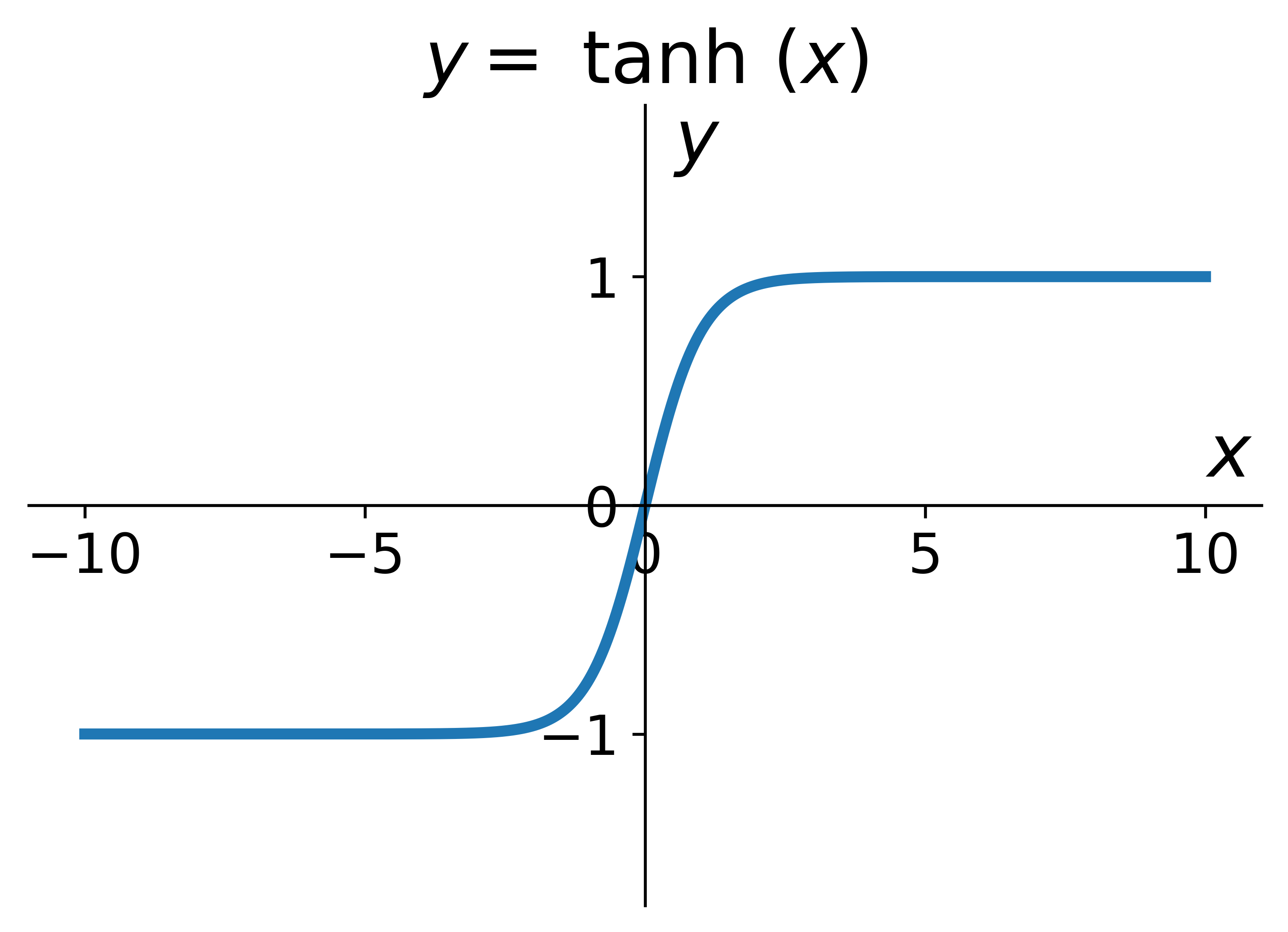}
    \includegraphics[width = 0.45\columnwidth]{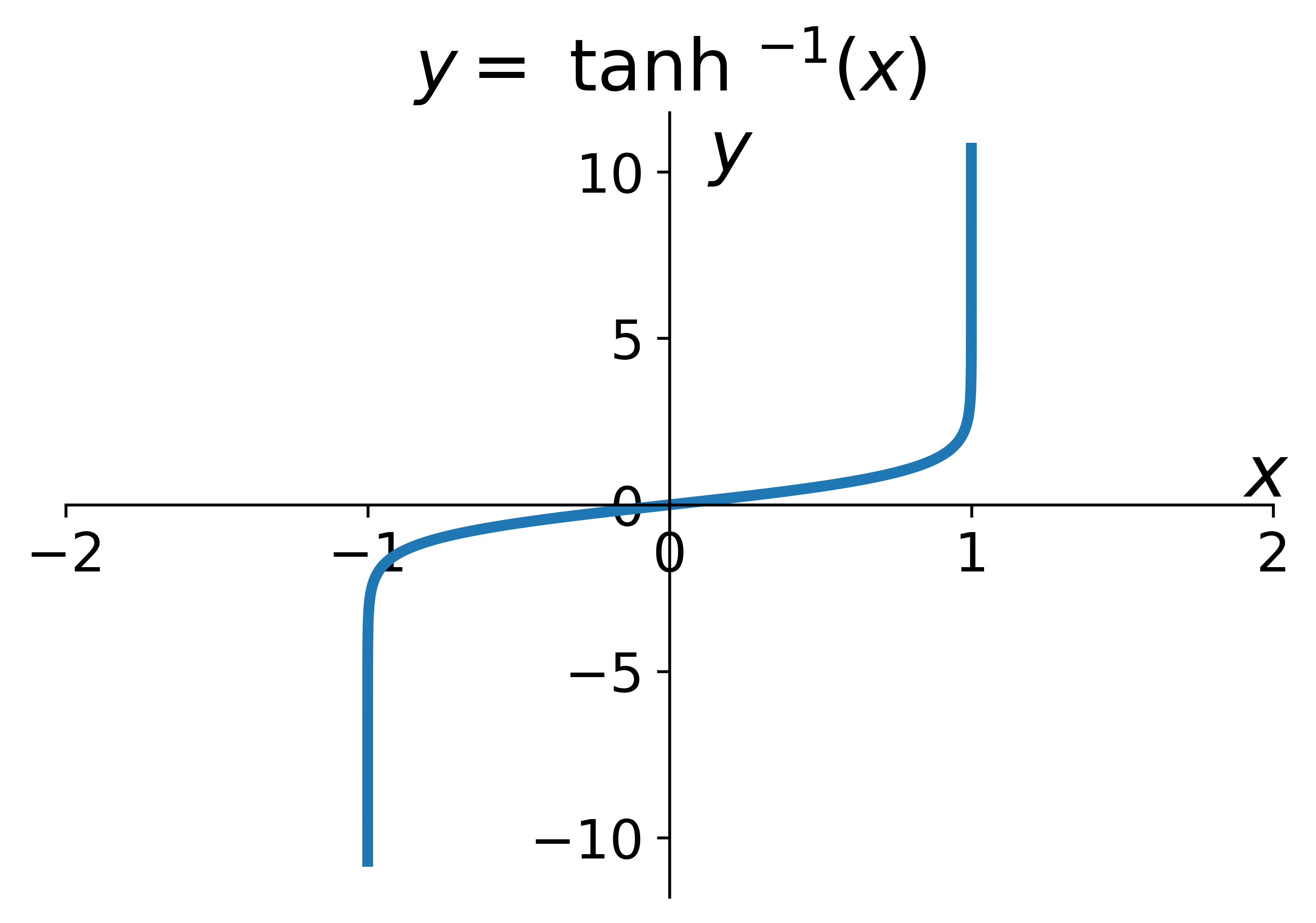}
    \caption{\textbf{Illustration of $\tanh(x)$ function (left) and $\tanh^{-1}(x)$ function (right).} When $|x|$ is close to 1, $|\tanh^{-1}(x)|$ approaches infinity.}
    \label{fig:tanh}
\end{figure}

ReLie~\cite{falorsi2019reparameterizing} performs normalizing flows on the Lie algebra of Lie groups, where for rotations in $\SO$, the Lie algebra is the axis-angle representation ($\theta\mathbf{e}$) in $\mathbb{R}^3$. ReLie applies normalizing flows in Euclidean space and then maps the Euclidean samples back to $\SO$ space through the exponential map. Noticing that the axis-angle representation periodically covers $\SO$ space, i.e., $(\theta+2i\pi)\mathbf{e}$, $i\in \mathbb{Z}$ represent the same rotation, ReLie restricts the output of the Euclidean normalizing flows in a sphere with the radius $r$ by an $r\cdot \operatorname{tanh}(\cdot)$ operation to resolve the infinity-to-one issue. However, ReLie still exhibits several drawbacks. Firstly, even if limited in a sphere with the radius of $\pi$, axis-angle representation is not a diffeomorphism to $\SO$ \cite{Zhou2019Continuity}, thus the normalizing flows cannot build diffeomorphic mappings between Lie algebra and Lie group and suffer from discontinuous rotation representation. Secondly, with the non-linear $\operatorname{tanh}(\cdot)$ operation, at the inverse stage, $\operatorname{tanh^{-1}}(\cdot)$ can yields infinitely large values, resulting in numerical instability (see Figure \ref{fig:tanh}). Note that this issue can not be solved by replacing $\operatorname{tanh}(\cdot)$ by other functions or varying the value $r$, since a non-linear mapping function is always needed to restrict $\mathbb{R}^3$ into a sphere.

In our experiment, ReLie fails to fit the \textit{peak} distribution (see Table 1 in the main paper). This is because the normalizing flows learn to push mostly all the points to the peak, and in the inverse process, points that are not close to the peak are pushed back near to the surface of the $r$-sphere in Lie algebra, which yields $\mathtt{NAN}$ in training and breaks the process.

\subsection{ProHMR}
ProHMR~\cite{kolotouros2021probabilistic} leverages 6D rotation representation and considers the first two columns of the rotation matrix as a 6D Euclidean vector. Thus, it applies Euclidean normalizing flows on the 6D vectors and maps the samples back to $\SO$ by Gram-Schmidt orthogonalization. Without any constraint, the infinity-to-one mapping clearly makes the probability density of a given rotation intractable. This is because a rotation in $\SO$ corresponds to infinity points in Euclidean space and the PDF of the rotation should be the integration of all the corresponding Euclidean points. Due to the unavailability of the probability densities, we do not incorporate it as our baseline in experiments.

\section{Application: Entropy Estimation of Arbitrary Distribution}

One outstanding feature of our Normalizing Flows compared to other probability inference methods on $\SO$ (like \cite{murphy2021implicit}) is its ability for efficient samples. \cite{murphy2021implicit} can only sample by querying a large number of rotations and calculating the probability density function. For highly peaked distribution, this method may fail as it is hard to have queries that are enough close to the peak such that the probability is not close to zero. However, we can sample by transforming $z$ sampled from a base distribution 
through our flows. Efficient sampling makes it possible to estimate properties of data $x$, for example, \textit{entropy} can be estimated via Monte Carlo:
\begin{equation}
 S = E[\log p(x)].
\end{equation}

\begin{figure*}[t]
    \centering
    \includegraphics[width=\linewidth]{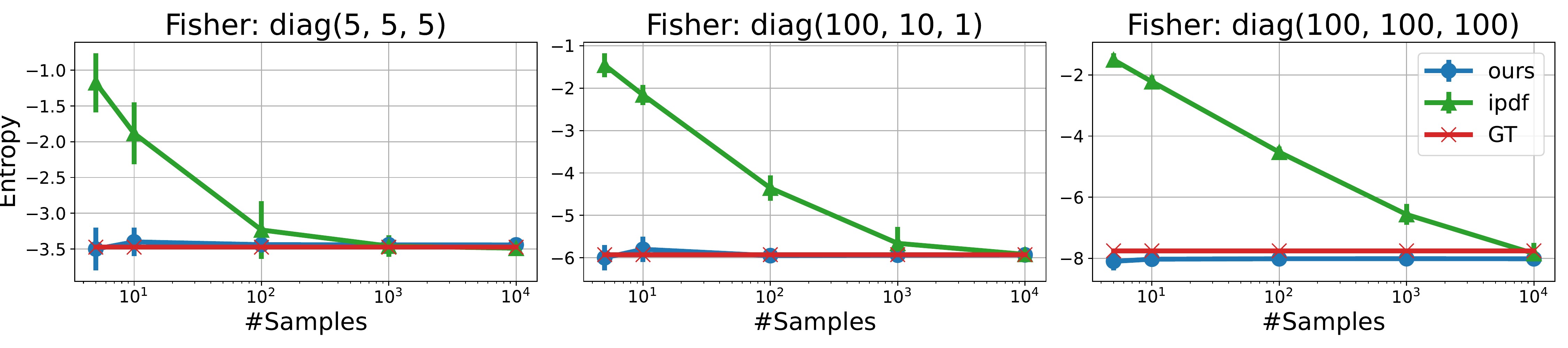}
    \caption{\textbf{Results of entropy estimation for three different target distributions.} We compare the mean and variance of estimated entropy after 100 times of sampling. We use uniform distribution as base distribution. Note that the horizontal axis is in log scale.
    }
    \label{fig:entropy}
\end{figure*}

In this experiment, we
compare our rotation NFs with Implicit-PDF in approximating the entropy of the target distributions.
In order to obtain the ground truth entropy for evaluation, we adopt multiple matrix Fisher distributions (whose entropy can be analytically computed) with different parameters as the target distributions. 
We sample 600k points from each target distribution as the training data and evaluate both our method and Implicit-PDF by randomly sampling N (N=5, 10, 100, 1k, 10k) points from the learned distributions.
The results are shown in Figure \ref{fig:entropy}. We can see that even when the sampling size is small, our rotation normalizing flows still achieve accurate estimation of entropy for different target distributions, while Implicit PDF fails to do so.

\par

\section{Calculations of Our Normalizing Flows}
\label{sec:calculation}
\subsection{Determinate of Jacobian}
\paragraph{Mobius Transformation}
The forward pass of Mobius transformation is defined as follows:
\begin{equation}
    c^\prime = f_\omega(c)=\frac{1-\Vert\omega\Vert^2}{\Vert c-\omega\Vert ^2}(c-\omega) -\omega.
\end{equation}
Where $c, c^\prime$ are vectors $\in\mathcal{S}^2$, $\omega$ is a point $\in \mathbb{R}^{3}$ that satisfies $\Vert\omega\Vert < \frac{\sqrt{2}}{2}$.The Jacobian matrix is given by partial derivatives of $c^\prime$ with respect to $c$,
\begin{equation}
    \frac{\mathrm{d}c^\prime}{\mathrm{d}c} = \frac{1-\Vert \omega\Vert^2}{\Vert c-\omega\Vert^2} (I - 2 \frac{(c - \omega)^T(c-\omega)}{\Vert c - \omega\Vert ^ 2})
\end{equation}
where $I$ is the 3$\times3$ identity matrix. \par
In our implementation, the Mobius transformation is a one-degree-of-freedom mapping, which maps unit vectors lying in the plane vertical to the unchanged condition column to unit vectors in the same plane. As any three-dimensional vector $c$ on the circle can be parameterized by an angle $\theta$ to reference vector $c_2$ where $c_2$ and $c_3$ are one pair of the orthogonal basis of the plane.
\begin{gather}
    c = \cos\theta c_2 + \sin\theta c_3\\
    c^\prime = \cos\theta^\prime c_2 + \sin\theta^\prime c_3
\end{gather}
\par
Via a change of variable formula, the determinant of Jacobian can be calculated as:
\begin{equation}
    J=\frac{\mathrm{d}\theta^\prime}{\mathrm{d}\theta} = \frac{\mathrm{d}\theta^\prime}{\mathrm{d}c^\prime}\frac{\mathrm{d}c^\prime}{\mathrm{d}\theta}
\end{equation}
Given that $c, c^\prime, c_2, c_3$ are unit vectors, the absolute value of determinant of Jacobian is thus calculated as:
\begin{equation}
|\det J|=|\frac{\mathrm{d}\theta^\prime}{\mathrm{d}\theta}|=\Vert\frac{\mathrm{d}c^\prime}{\mathrm{d}\theta}\Vert
\end{equation}
Where $\frac{\mathrm{d}c^\prime}{\mathrm{d}\theta}$ are given by:
\begin{gather}
    \frac{\mathrm{d}c^\prime}{\mathrm{d}\theta}=\frac{\mathrm{d}c^\prime}{\mathrm{d}c}\frac{\mathrm{d}c}{\mathrm{d}\theta}\\
    \frac{\mathrm{d}c}{\mathrm{d}\theta}= - \sin\theta c_2 + \cos\theta c_3 
\end{gather}

\paragraph{Affine Transformation}
The forward of quaternion affine transformation is given by: 
\begin{equation}
    \mathbf{q'}=g(\mathbf{q}) = \frac{W\mathbf{q}}{\Vert W\mathbf{q}\Vert}
\end{equation}
The determinate of Jacobian of affine transformation is very straightforward and is given by:
\begin{equation}
    \det{J(\mathbf{q})} = \frac{\det{W}}{\Vert W\mathbf{q}\Vert^4}
\end{equation}

\subsection{Inverse}
\paragraph{Mobius Transformation}
The inverse of Mobius transformation is implemented by connecting $-c^\prime$ and parameters $\omega$ with a straight line that intersects with $S^D$ at $c$. The explicit expression for the inverse of Mobius transformation is given as follows: 
\begin{equation}
    f^{-1}_\omega(c^\prime)=\frac{1-\Vert \omega\Vert^2}{\Vert c^\prime+ \omega\Vert^2}(c^\prime+\omega)+\omega
\end{equation}
The forward process and the inverse process have the same computational complexity. 
\par
However, as we utilize a linear combination, i.e. the weighted sum of angles to improve the expressivity of Mobius transformation, there is no analytical inverse for combined Mobius transformation, so we use binary search algorithms to find its inverse, as the combined angle $\theta^\prime$ are constrained within$(-\pi/2, \pi/2)$ and there's no discontinuity around the boundary $\pm \pi/2$. The computational complexity for inversing the Mobius transformation is $O(\log(1/\epsilon)$, where $\epsilon$ is the computational error. 

\paragraph{Affine Transformation}
The inverse of affine transformation is implemented as:
\begin{equation}
    g^{-1}(\mathbf{q^\prime})=\frac{W^{-1}\mathbf{q^\prime}}{\Vert W^{-1}\mathbf{q^\prime}\Vert}\label{eq:affine-supp}
\end{equation}
\par
Where $W$ is a $4\times 4$ invertible matrix and $\mathbf{q}^\prime$ is a quaternion. We find that the inverse and forward processes of affine transformation have the same expression form while the invertible matrix $W$ in the forward path is replaced by its inverse matrix $W^{-1}$ in the inverse path. We attribute this feature to the geometry of affine transformation.
\par
Via SVD decomposition, the affine transformation is divided into two types of operations: 1)rotation, 2) scaling and normalization. The inverse of rotation operation $R$ takes place by multiplying $R^{-1}$ which is straightforward, whereas the inverse of scaling and normalization is also implemented by taking the inverse of $S$ which sounds non-trivial. 
\par
\begin{figure}
    \centering
    \includegraphics[width = 0.7\columnwidth]{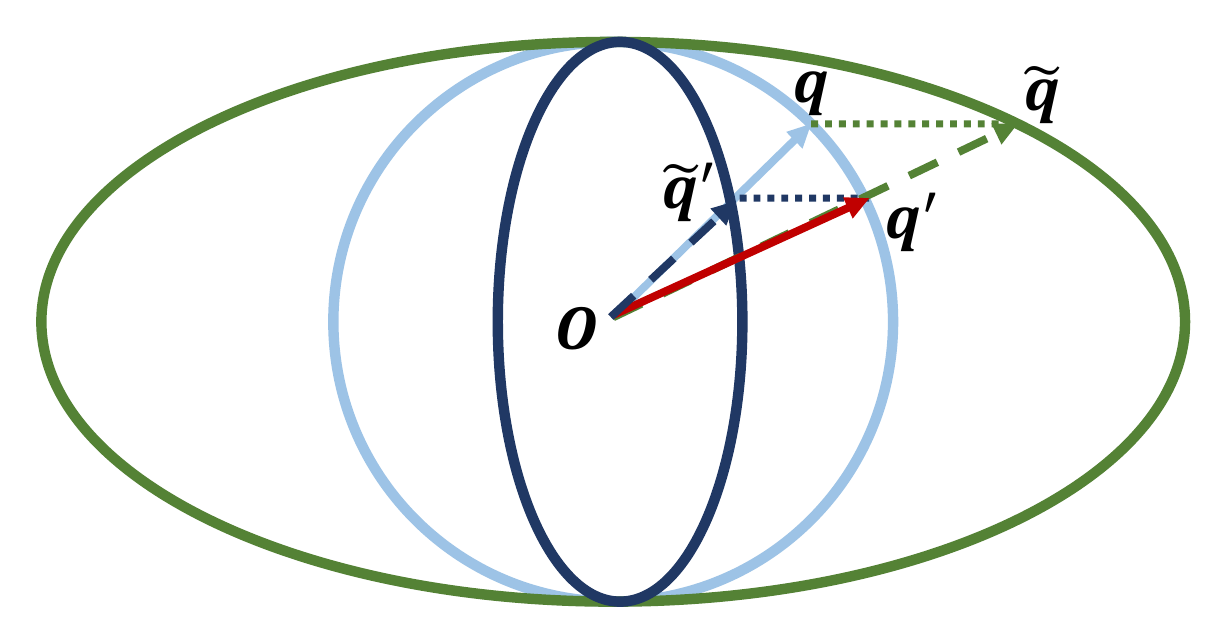}
    \caption{Forward and inverse of scaling and normalization operation in quaternion affine transformation. $\mathbf{q}$ are scaled to $\tilde{\mathbf{q}}$ by multiplying $s_1, s_2, s_3, s_4$ on its coordinates and then normalized to the unit sphere $\mathbf{q^\prime}$. Inversely, $\mathbf{q^\prime}$ are scaled to $\tilde{\mathbf{q}}^\prime$ by $1/s_1, 1/s_2, 1/s_3, 1/s_4$ and normalized to $\mathbf{q}$. $\Delta O\tilde{\mathbf{q}}^\prime\mathbf{q^\prime}$ and $\Delta O\mathbf{q}\tilde{\mathbf{q}}$ are similar triangles.}
    \label{fig:affine-supp}
\end{figure}

The feature is caused by a geometry coincidence, as illustrated in Figure \ref{fig:affine-supp}. $\mathbf{q}$ are scaled to $\tilde{\mathbf{q}}$ by multiplying $s_1, s_2, s_3, s_4$ on its coordinates and then normalized to the unit sphere $\mathbf{q^\prime}$. Inversely, $\mathbf{q^\prime}$ are scaled to $\tilde{\mathbf{q}}^\prime$ by $1/s_1, 1/s_2, 1/s_3, 1/s_4$ and normalized to points intersects with $S^3$. $\Delta O\tilde{\mathbf{q}}^\prime\mathbf{q^\prime}$ and $\Delta O\mathbf{q}\tilde{\mathbf{q}}$ are similar triangles as:
\begin{equation}
    \frac{\overline{O\tilde{\mathbf{q}}^\prime}}{\overline{O\mathbf{q}}} = \frac{\overline{O\mathbf{q^\prime}}}{\overline{O\tilde{\mathbf{q}}}} = \frac{\overline{\tilde{\mathbf{q}}^\prime\mathbf{q}^\prime}}{\overline{\mathbf{q}\tilde{\mathbf{q}}}}
\end{equation}
As $O, \mathbf{q}^\prime, \tilde{\mathbf{q}}$ are on the same line and $\angle \tilde{\mathbf{q}}^\prime O \mathbf{q}^\prime = \angle\mathbf{q} O \tilde{\mathbf{q}}$ due to the property of similar triangles, $O, \tilde{\mathbf{q}}^\prime, \textbf{q}$ are on the same line, thus normalized points $\tilde{\mathbf{q}}^\prime$ are $\mathbf{q}$, which is the inverse of affine transformation.
\section{ModelNet10-SO(3) and Pascal3D+ detailed results}
Table \ref{tab:modelnet-all} and Table \ref{tab:pascal-all} show per-category metrics for ModelNet10-SO(3) and Pascal3D+ dataset respectively. Note that our method achieves significantly lower median errors in experiments on ModelNet10-SO3 in all of the categories except the \textit{batutub} category, where \textit{batutub} images are well-known to exhibit severe symmetry and all the methods have unreasonably poor performance.

\begin{table*}[ht]
  \centering
  \resizebox{0.9\textwidth}{!}{
    \begin{tabular}{@{}llccccccccccc}
      \toprule
       &                                      & {avg.}     & {bathtub}  & {bed}      & {chair}    & {desk}     & {dresser}  & {tv}       & {n. stand} & {sofa}     & {table}    & {toilet}   \\
      \midrule
      
      \multirow{4}{*}{Acc@15\textdegree}
    & Deng et al.\cite{deng2022deep}            
    & 0.562      & 0.140      & 0.788      & 0.800      & 0.345      & 0.563      & 0.708      & 0.279      & 0.733      & 0.440      & 0.832      \\      
    & Prokudin et al.\cite{prokudin2018deep}        
    & 0.456      & 0.114      & 0.822      & 0.662      & 0.023      & 0.406      & 0.704      & 0.187      & 0.590      & 0.108      &  0.946 \\
    & Mohlin et al.\cite{mohlin2020probabilistic} 
    &  0.693 &  0.322 & 0.882 & 0.881 &  0.536 &  0.682 &  0.790 &  0.516 & 0.919 &  0.446 & 0.957 \\      
    & IPDF \cite{murphy2021implicit}                          & 
    0.719 & 0.392 & 0.877 &  0.874 & 0.615 & 0.687 & 0.799 & 0.567 &  0.914 & 0.523 & 0.945      \\

    & Ours(Uni.) & \textbf{0.760} &	\underline{0.402} & \textbf{0.896} & \textbf{0.927}	& \textbf{0.704} & \textbf{0.753} & \textbf{0.843} & \textbf{0.602} & \textbf{0.939} & \textbf{0.561}	& \underline{0.975}\\
    & Ours(Fisher) & \underline{0.744} & \textbf{0.439} & \underline{0.890} & \underline{0.909} & \underline{0.638} & \underline{0.715} & \underline{0.810} & \underline{0.585} & \underline{0.938} & \underline{0.535} & \textbf{0.978}
    \\

      \midrule
      \multirow{4}{*}{Acc@30\textdegree}                   
    & Deng et al.\cite{deng2022deep}            & 0.694      & 0.325      & 0.880      & 0.908      & 0.556      & 0.649      & 0.807      & 0.466      & 0.902      & 0.485      & 0.958      \\      
    & Prokudin et al.\cite{prokudin2018deep}        & 0.528      & 0.175      & 0.847      & 0.777      & 0.061      & 0.500      & 0.788      & 0.306      & 0.673      & 0.183      & 0.972 \\
    & Mohlin et al.\cite{mohlin2020probabilistic} 
    & 0.757& 0.403 & \textbf{0.908} & 0.935 & 0.674 & \underline{0.739} & \underline{0.863} & 0.614 & \underline{0.944} &  0.511 & 0.981 \\      
    & IPDF \cite{murphy2021implicit}                         
    & 0.735 & 0.410 &  0.883 &  0.917 &  0.629 &  0.688 &  0.832 &  0.570 &  0.921 & 0.531 & 0.967      \\
    & Ours(Uni.) & \textbf{0.774} &	\underline{0.419}	& \underline{0.904} & \textbf{0.946} & \textbf{0.722} & \textbf{0.766} & \textbf{0.868}	& \textbf{0.617} & \textbf{0.948}	& \textbf{0.567} & \underline{0.982}\\
    & Ours(Fisher) & \underline{0.768} & \textbf{0.460} & 0.898 & \underline{0.934} & \underline{0.694} & 0.738 & 0.859 & \underline{0.615} & \textbf{0.948} & \underline{0.544} & \textbf{0.987}
    \\

      \midrule
      \multirow{4}{*}{\shortstack{Median \\ Error ($^\circ$)}} 
    & Deng et al.\cite{deng2022deep}            
    & 32.6       & 147.8      & 9.2        & 8.3        & 25.0       & 11.9       & 9.8        & 36.9       & 10.0       & 58.6       & 8.5        \\
    & Prokudin et al.\cite{prokudin2018deep}        
    & 49.3       & 122.8 & 3.6   & 9.6        & 117.2      & 29.9       & 6.7        & 73.0       & 10.4       & 115.5      &  4.1   \\
    & Mohlin et al.\cite{mohlin2020probabilistic} 
    & 17.1  & \textbf{89.1}  &  4.4   & 5.2   & 13.0  &  6.3   &  5.8   &  13.5  & 4.0   &  25.8  & 4.0   \\      
    & IPDF \cite{murphy2021implicit}                         
    & 21.5  & 161.0      & 4.4   & 5.5   & 7.1   & 5.5   & 5.7   & 7.5   &  4.1   & 9.0   & 4.8        \\
    & Ours(Uni.) & \textbf{14.6}	& 124.8	& \textbf{1.5} & \textbf{2.8} & \textbf{2.7} &	\textbf{1.5} &\textbf{2.6} &	\textbf{2.4} & \textbf{1.5}	& \textbf{3.9} & \textbf{2.0}\\
    & Ours(Fisher) & \textbf{12.2} & \underline{91.6} & \underline{1.8} & \underline{3.0} & \underline{5.5} & \underline{2.0} & \underline{3.2} & 4.3 & \underline{1.6} & \underline{6.7} & \underline{2.1}\\

      \bottomrule
    \end{tabular}}
  \caption{\textbf{Numerical results of rotation regression on ModelNet10-SO3 dataset.} We adopt 15$^{\circ}$ accuaracy, $30^{\circ}$ accuaracy and median error as the evaluation metrics. The best performance is shown in \textbf{bold} and the second best is with \underline{underlined}.}
  \label{tab:modelnet-all}
\end{table*}
\begin{table*}[ht]
  \centering
  \resizebox{0.9\textwidth}{!}{
    \begin{tabular}{@{}llccccccccccccc}
      \toprule
       &                                      & {avg.}     & {aero}  & {bike}      & {boat}    & {bottle}     & {bus}  & {car}       & {chair}      & {table}    & {mbike} & {sofa} & {train} & {tv} \\
      \midrule
    
      \multirow{4}{*}{Acc@30\textdegree}                   
    & Liao et al.\cite{liao2019spherical}            &0.819 & 0.82 & 0.77 & 0.55 & 0.93 & 0.95 & 0.94 & 0.85 & 0.61 & 0.80 & \underline{0.95} & \textbf{0.83} & 0.82  \\      
    
    & Mohlin et al.\cite{mohlin2020probabilistic} 
    & 0.825 & 0.90 & \underline{0.85} & 0.57 & 0.94 & 0.95 & \textbf{0.96} & 0.78 & 0.62 & 0.87 & 0.85 & 0.77 & 0.84 \\      
    & Prokudin et al.\cite{prokudin2018deep}      & 0.838 & \textbf{0.89} & 0.83 & 0.46 & \textbf{0.96} & 0.93 & 0.90  & 0.80 & \underline{0.76} & \underline{0.90} & 0.90 & \underline{0.82} & \textbf{0.91} \\
    & Tulsiani \& Malik \cite{tulsiani2015viewpoints} & 0.808 & 0.81 & 0.77 & \underline{0.59} & 0.93 & \textbf{0.98} & 0.89 & 0.80 & 0.62 & 0.88 & 0.82 & 0.80 & 0.80 \\
    & Mahendran et al. \cite{mahendran2018mixed} & \underline{0.859} & \underline{0.87} & 0.81 & \textbf{0.64} & \textbf{0.96} & \underline{0.97} & \underline{0.95} & \underline{0.92} & 0.67 & 0.85 & \textbf{0.97} & \underline{0.82} & \underline{0.88} \\
    & IPDF \cite{murphy2021implicit}                         
    & 0.837 & 0.81 & \underline{0.85} & 0.56 & 0.93 & 0.95 & 0.94 & 0.87 & \textbf{0.78} & 0.85 & 0.88 & 0.78 & 0.86      \\
    & Ours(Uni.)
    & 0.827 & 0.83 & 0.78 & 0.56 & \underline{0.95} & 0.96 & 0.93 & 0.87 & 0.62 & 0.85 & 0.90 & 0.81 & 0.86 \\
    & Ours(Fisher)
    & \textbf{0.863} & \textbf{0.89} & \textbf{0.89} & 0.55 & \textbf{0.96} & \textbf{0.98} & \underline{0.95} & \textbf{0.94} & 0.67 & \textbf{0.91} & \underline{0.95} & \underline{0.82} & 0.85 \\

      \midrule
      \multirow{4}{*}{\shortstack{Median \\ Error ($^\circ$)}} 
    & Liao et al.\cite{liao2019spherical}            
    & 13.0 & 13.0 & 16.4 & 29.1 & 10.3 & 4.8 & 6.8 & 11.6 & 12.0 & 17.1 & 12.3 & 8.6 & 14.3        \\
    & Mohlin et al.\cite{mohlin2020probabilistic} 
    & 11.5 & 10.1 & 15.6 & 24.3 & 7.8 & 3.3 & 5.3 & 13.5 & 12.5 & 12.9 & 13.8 & 7.4 & 11.7   \\  
    & Prokudin et al.\cite{prokudin2018deep}        
    & 12.2 & 9.7 & 15.5 & 45.6 & \textbf{5.4} & 2.9 & \underline{4.5} & 13.1 & 12.6 & \textbf{11.8} & \underline{9.1} & \textbf{4.3} & 12.0   \\
    & Tulsiani \& Malik \cite{tulsiani2015viewpoints} & 13.6 & 13.8 & 17.7 & \underline{21.3} & 12.9 & 5.8 & 9.1 & 14.8 & 15.2 & 14.7 & 13.7 & 8.7 & 15.4 \\
    & Mahendran et al. \cite{mahendran2018mixed} & \underline{10.1} & \textbf{8.5} & 14.8 & \textbf{20.5} & 7.0 & 3.1 & 5.1 & 9.3 & 11.3 & 14.2 & 10.2 & \underline{5.6} & 11.7 \\
    & IPDF \cite{murphy2021implicit}                         
    & 10.3 & 10.8 & \underline{12.9} & 23.4 & 8.8 & 3.4 & 5.3 & 10.0 & \textbf{7.3} & 13.6 & 9.5 & 6.4 & 12.3        \\
    & Ours(Uni.)
    & 10.2 & \underline{8.9} & 15.2 & 24.9 & \underline{6.9} & \textbf{2.9} & \textbf{4.3} & \textbf{8.7} & 10.7 & \underline{12.8} & 9.3 & 6.3 & \textbf{11.3}  \\
    & Ours(Fisher)
    & \textbf{9.9} & 9.6 & \textbf{12.4} & 22.7 & 7.5 & \underline{3.1} & 4.8 & \underline{9.2} & \underline{8.6} & 13.5 & \textbf{8.6} & 6.7 & \underline{11.6} \\

      \bottomrule
    \end{tabular}}
  \caption{\textbf{Numerical results of rotation regression on Pascal3D+ dataset.} We adopt $30^{\circ}$ accuaracy and median error as the evaluation metrics. The best performance is shown in \textbf{bold} and the second best is with \underline{underlined}.}
  \label{tab:pascal-all}
\end{table*}

\section{Analysis of Evaluation Methods}
\subsection{Log Likelihood Evaluation}
In case of main experiments (rotation distribution fitting) and SYMSOL I/II datasets, we exploit \textbf{log likelihood} averaged over test data or test set annotations as evaluation metrics. It is implemented by transforming data through the flow to its base distribution and predicting log-likelihood.
\subsection{Sample Generation}
Sometimes pose estimation tasks require single-value prediction. Estimated rotations are generated via flowing samples from base distribution inversely and obtained a set of samples following target distribution. We pick the sample with the highest probability likelihood as our single prediction.

In SYMSOL I datasets, we present \textbf{\textit{spread}} following IPDF\cite{murphy2021implicit}. Given a complete set of equivalent ground truth rotations, \textit{spread} is defined as expectation of angular deviation to any of the ground truth values: $\mathbb{E}_{R\sim p(R|x)}[\min_{R^\prime\in\{R_{GT}\}}d(R, R^\prime)]$, where $x$ is the given image and $d(R, R^\prime)$ is the relative angles between $R$ and $R^\prime$. This measures how close the samples are to the ground truth poses. 
It is calculated as the mean of the relative angle between each of the generated samples and the ground truth poses closest to it.

In ModelNet10-SO(3) and Pascal3D+ datasets, we calculate the angular error of our single-value rotation prediction with single ground truth value, and report \textbf{error median}, accuracy at threshold 15$^\circ$ (for ModelNet-SO(3) only), 30$^\circ$, i.e. \textbf{Acc@15, Acc@30}.

\subsection{Ablations of other sample generation methods}
We compare our inference method to those used in IPDF \cite{murphy2021implicit}, which is implemented via first evaluating a grid $\{R_i\}$ or samples randomly sampled on $\SO$, and solving
\begin{equation}
    R_x^*={\arg \max}_{R\in \SO} p(R|x)
    \label{eq: gradient}
\end{equation}
to pick the single pose prediction (We called the method \textit{PDF} for short). The prediction can be made more accurate with gradient ascent of Equation (\ref{eq: gradient}) (\textit{Grad} for short). The equivolumetric grids are generated first generating equal area grids on the 2-sphere with HEALPix method, and cover $\SO$ with Hopf fibration by sampling equivolumetric points on a circle with each point on the 2-sphere.

Table \ref{tab:evaluation} shows that with few numbers of samples ($\sim$ 5), our method is capable of capturing accurate results with little computational cost ($\sim$ 1-2 minutes for evaluating the whole test sets). While for PDF and Grad, they require high-resolution grids to obtain high accuracy and the computation cost is not affordable when evaluating a large dataset.

\begin{table*}[!h]
\centering
\setlength{\tabcolsep}{3pt}

\caption{
\textbf{Abalations on evaluation time}. We report results on conditional rotation regression tasks of ModelNet10-SO3. We compare the time required to generate estimated rotation via inversing the flow (Ours), inferringthe probability of a grid(PDF), gradient descent of probility of a grid(Grad). The number of samples used in PDF and Grad corespond to the HEALPix-SO(3) grids of levels 2, 3 and 4 respectively. 
}
\begin{tabular}
{lcccccc}
  \toprule

  Method & Number of samples & {evaluation time(total)$\downarrow$} & \multicolumn{1}{c}{evulation time(s)/iter$\downarrow$} & Acc@15$^\circ\uparrow$ & Acc@30$^\circ\uparrow$ & Med. ($^\circ$)$\downarrow$\\

  \midrule

  Ours & 1 & 1.2 min &  0.508 & 0.740 & 0.760 & 14.4\\
  Ours & 5 & 1.3 min& 0.525 & \underline{0.759} & \underline{0.773} & \underline{14.0}\\
  Ours & 50 & 1.4 min & 0.555 & \underline{0.759} & \underline{0.773} & 14.1\\
  Ours & 500 & 4.8 min & 1.97 & \textbf{0.760} & \underline{0.773} & 14.8 \\
  Ours & 5k & 0.7 h & 16.8 & \textbf{0.760} & \textbf{0.774} & 14.6\\
  \hline
  PDF & 4.6k & 0.6 h & 0.122 & 0.480 & 0.611 & 31.8\\
  PDF & 37k &  1.4 h &  0.289 & 0.699 & 0.745 & 18.2\\
  PDF & 295k & 9.5 h & 1.88 & 0.755 & 0.771 & 18.0\\
  \hline
  Grad & 4.6k & 2.2h &  0.446 & 0.496 & 0.612 & 31.6\\
  Grad & 37k & 3.0h &  0.610 & 0.704 & 0.748 & 16.6\\
  Grad & 295k & 11.0h &  2.20 & 0.758 & 0.772 & \textbf{13.8}\\
\bottomrule
\end{tabular}

\label{tab:evaluation}
\end{table*}

\section{Implementation Details}

We use Adam as our optimizer with a learning rate of 1e-4. 

\noindent{\textbf{Detailes of $\frac{\sqrt{2}}{2}$ trick}} In Mobius coupling layers, each $\omega$ is predicted by first predicting $\omega^{\prime}$ by a [64, 64, 64, 64] multi-layer perceptron (MLP), ReLU as activation function and a residual connection between the first and the last layer with the condition part as input, and then projecting $\omega^\prime$ to the vertical plane of condition column via Gram-Smith process. It is then reparameterized by $\omega = \frac{0.7}{{1+\Vert \omega\Vert}}\omega$ to constrain it within $\frac{\sqrt{2}}{2}$ sphere.

\noindent{\textbf{Unconditional Experiments}} we use 24 layers of blocks (24 Mobius + 24 Affine) with the combination of 64 Mobius transformations at a batch size of 64 for 50k steps. 

\noindent{\textbf{SYMSOL}} In experiments on SYMSOL, we add a conditional Affine transformation at the beginning of our flow (near the target distribution) and use 21 layers of blocks (64-combination conditional Mobius coupling layers + unconditional affine transformation) at a batch size of 128 for 900k steps. 
Conditional features with dimension 512 are captured by an ImageNet pre-trained ResNet50 following settings in IPDF\cite{murphy2021implicit}. We trained a single model for SYMSOL I for 5 categories and trained a single for each category in SYMSOL also following IPDF.

\noindent{\textbf{ModelNet10-SO3}} For experiments on ModelNet10-SO3, 24 layers of blocks (64-combination conditional Mobius coupling layers + conditional affine transformation) at a batch size of 128 for 250k steps are used. We also use the strategy of learning rate decay and multiply the learning rate by 0.1 at 30 and 40 epochs. Conditional features with dimension 2048 are captured by an ImageNet pre-trained ResNet101 following settings in Mohlin et al.\cite{mohlin2020probabilistic}. We train a single model for the whole dataset and use a 32-dimensional positional embedding for 10 categories.

\noindent{\textbf{Pascal3D+}}
 For Pascal3D+, 24 layers of blocks (64-combination conditional Mobius coupling layers + conditional affine transformation) at a batch size of 128 for 350k steps are used. We also use the strategy of learning rate decay and multiply the learning rate by 0.1 at 200k and 250k iterations when using uniform distribution as base distribution, 12 and 15 epochs when using pre-trained fisher as base distribution respectively. Conditional features with dimension 2048 are captured by an ImageNet pre-trained ResNet101. We train a single model for the whole dataset and use a 32-dimensional positional embedding for 10 categories. 



\end{document}